\documentclass{article}

\usepackage{amsmath} 
\usepackage{arxiv}

\usepackage[utf8]{inputenc} 
\usepackage[T1]{fontenc}    
\usepackage{hyperref}       
\usepackage{url}            
\usepackage{booktabs}       
\usepackage{amsfonts}       
\usepackage{nicefrac}       
\usepackage{microtype}      
\usepackage{lipsum}		
\usepackage{graphicx}
\usepackage{doi}

\usepackage{multirow}
\usepackage{rotating}
\usepackage{lscape}
\usepackage{soul}
\usepackage{graphicx}
\usepackage{float}
\usepackage{url}
\newcommand{\floor}[1]{\lfloor #1 \rfloor}

\title{A Recurrent YOLOv8-based framework for Event-Based Object Detection}

\author{
Diego A. Silva$^{1}$, Kamilya Smagulova$^{1}$, Ahmed Elsheikh$^{2}$, Mohammed E. Fouda$^{3,*}$, and Ahmed M. Eltawil$^{1}$ \\$^{1}$Electrical and Mathematical Sciences and Engineering Division, KAUST, Thuwal, Saudia Arabia. \\
$^{2}$Mathematics and Engineering Physics Dept., Faculty of Engineering, Cairo University, Giza, Egypt. \\
$^{3}$Compumacy for Artificial Intelligence solutions, Cairo, Egypt. \\
\texttt{*foudam@uci.edu} 
}

\begin{document}
\onecolumn


\maketitle

\begin{abstract}

Object detection is crucial in various cutting-edge applications, such as autonomous vehicles and advanced robotics systems, primarily relying on data from conventional frame-based RGB sensors. However, these sensors often struggle with issues like motion blur and poor performance in challenging lighting conditions. In response to these challenges, event-based cameras have emerged as an innovative paradigm. These cameras, mimicking the human eye, demonstrate superior performance in environments with fast motion and extreme lighting conditions while consuming less power. This study introduces Recurrent YOLOv8 (ReYOLOV8), an advanced object detection framework that enhances a leading frame-based detection system with spatiotemporal modeling capabilities. We implemented a low-latency, memory-efficient method for encoding event data to boost the system's performance. Additionally, we developed a novel data augmentation technique tailored to leverage the unique attributes of event data, thus improving detection accuracy. Our framework underwent evaluation using the comprehensive event-based datasets Prophesee’s Generation 1 (GEN1) and Person Detection for Robotics (PEDRo). Our models outperformed all comparable approaches in the GEN1 dataset, focusing on automotive applications, achieving mean Average Precision (mAP) improvements of 5\%, 2.8\%, and 2.5\% across nano, small, and medium scales, respectively. These enhancements were achieved while reducing the number of trainable parameters by an average of 4.43\% and maintaining real-time processing speeds between 9.2ms and 15.5ms. On the PEDRo dataset, which targets robotics applications, our models showed mAP improvements ranging from 9\% to 18\%, with 14.5x and 3.8x smaller models and an average speed enhancement of 1.67x. These results highlight the transformative potential of integrating event-based technologies with advanced object detection frameworks, paving the way for more robust and efficient visual processing systems in dynamic and challenging environments.
\end{abstract}


\section{Introduction}

Object Detection involves the dual processes of locating and categorizing objects within an image, serving as a critical function in a multitude of fields, including Autonomous Driving \cite{object_detection_autonomous_driving}, Robotics \cite{object_detection_robotics}, and Surveillance \cite{object_detection_surveillance}. Deep Learning algorithms primarily drive advancements in this area, with the You Only Look Once (YOLO) detectors, based on Convolutional Neural Networks (CNN), emerging as a prominent choice in academia and industry. Renowned for its real-time capabilities, YOLO stands out for its efficient performance with minimal parameters, as documented in various studies \cite{survey_obj_det_deep_learning}. Over time, YOLO has undergone iterations and enhancements, making it faster and more robust in handling object detection tasks \cite{survey_yolo}.

In contemporary computer vision applications, the standard practice involves processing images captured by cameras that detect light in red, green, and blue wavelengths (RGB). While modern RGB sensors excel in providing high-resolution and detailed frames, they are susceptible to motion blur during high-speed movements, and their limited High Dynamic Range (HDR) poses challenges in complex lighting scenarios \cite{neuromorphic_ads}. In contrast, event-based cameras operate based on changes in illumination rather than absolute light levels, drawing inspiration from the human eye's visual data processing mechanism. This novel approach results in sparse data sequences comprising spatial coordinates, timestamps, and polarity information triggered by variations in light stimuli at specific pixels \cite{posch2014retinomorphic}. Event-based cameras offer distinctive advantages, including ultra-low latency in the microsecond range, an HDR typically exceeding 120dB, and power consumption in the milliwatt range. These characteristics render event-based cameras particularly suitable for time-critical tasks and challenging lighting conditions, making them a preferred choice in applications where swift responsiveness and adaptability to varying light environments are paramount \cite{eventbasedsurvey}.

The majority of existing object detectors tailored for event-based data primarily target autonomous driving scenarios, typically relying on datasets like Prophesee's Generation 1 (GEN1) \cite{gen1} and 1 MegaPixel \cite{red_megapixel}, alongside robotics applications supported by the recently introduced Person Detection in Robotics (PEDRo) dataset \cite{pedro}. These detectors are commonly categorized into two groups based on their approach to event stream handling: direct processing of sparse event streams and densification before processing. The former group includes Graph Neural Networks (GNNs) \cite{graph_event_first}, Spiking Neural Networks (SNNs) \cite{spikingDenseNet}, and sparse CNNs \cite{asynet}. On the other hand, detectors in the second group first densify events before applying feature extractors such as CNNs \cite{red_megapixel} and transformers \cite{rvt} and occasionally incorporate Recurrent Neural Networks (RNNs) or State Space Models (SSM) \cite{event-ssm} for modeling temporal relationships. On both categories, the feature extractors are combined with detection heads commonly found in frame-based detection models, such as the YOLO family \cite{rvt}, \cite{get}, \cite{ergo12}, RetinaNet \cite{retinanet}, and Single-Shot Detector (SSD) \cite{ssd}, which are proven to provide good detection capability in event-based scenarios.

Despite recent advancements in GNNs and SNNs, detectors relying on densified event representations consistently outperform them by a significant margin, as evidenced in studies such as those by \cite{dagr}, \cite{easnn}, and \cite{ergo12}. Notably, top-performing detectors often leverage state-of-the-art detection heads borrowed from frame-based literature, leading to superior performance. Considering this trend, this research has opted to build upon the YOLOv8 framework \cite{yolov8} as the foundation for the event-based detector development. YOLOv8's exceptional performance and real-time processing capabilities make it a compelling choice over other alternatives, with a focus on achieving enhanced performance with reduced parameters. Recognizing the success of integrating spatial feature extractors and RNNs in event data processing, as demonstrated in works like \cite{red_megapixel}, \cite{astmnet}, and \cite{rvt}, the YOLOv8 framework has been enhanced to incorporate Convolutional Long-Short Term Memory (ConvLSTM) cells, along with the implementation of Truncated-Backpropagation Through Time (T-BPTT) for RNN training \cite{convlstm}.

A critical aspect of current event-based object detectors lies in the dense encodings used for effectively feeding event stream data into neural networks. Various encoding strategies have been proposed in the literature, each aiming to retain crucial information from event streams. These strategies range from simple projections on 2D planes based on event counting \cite{event_frame}, \cite{eventCount1}, timestamp manipulation techniques \cite{sae}, \cite{hats}, \cite{inceptive}, to hybrid approaches combining both methods \cite{eventCount2}, \cite{cstr}. Other methodologies involve segmenting streams into spatiotemporal volumes \cite{voxelgrid}, \cite{rvt}, \cite{mdes}, \cite{aec}, \cite{dstdnet}, while learning-based encodings \cite{est}, \cite{dmanet}, Bayesian optimization \cite{ergo12}, and the utilization of First-In-First-Out (FIFO) buffers \cite{tore}, \cite{aed} have also been proposed. Each of these approaches presents distinct trade-offs, impacting detection performance, encoding latency, and memory requirements associated with event inputs. While existing literature assesses the impact of these encoding choices on detection performance, inference time, and event processing rates, a comprehensive system-level perspective is lacking. To address this gap and introduce a memory-efficient fast encoding alternative, this work introduces a novel event representation titled Volume of Ternary Event Images (VTEI). A comparative analysis against closely related alternatives was conducted, not only focusing on parameters like latency but also assessing hardware-related factors such as data unit encoding size, memory footprint under various event rates, compression ratios, and bandwidth requirements.

Moreover, recognizing the lack of techniques that specifically address event-based feature augmentation, this study introduces a novel approach known as Random Polarity Suppression (RPS). This method involves randomly suppressing all events associated with a particular polarity, enabling the detector to learn object-relevant features in a polarity-agnostic manner and mitigating potential biases in polarity distribution that could exist within the training dataset.

In summary, the contributions of this work are:

\begin{itemize}
    \item A Recurrent-Convolutional Event-Based Object Detection network was introduced by means of the modification of the well-acknowledged real-time detector YOLOv8. The resulting framework, called Recurrent YOLOv8 (ReYOLOv8), was based on the addition of recurrent blocks and training with T-BPTT to the original framework, turning it capable of performing long-range spatiotemporal modeling;

    \item A fast and lightweight memory encoding called Volume of Ternary Event Images (VTEI) was proposed. This format is capable of retaining temporal information from event streams while presenting low latency, low bandwidth, high sparsity, and high compression ratio;

    \item A novel data augmentation technique based on Random Polarity Suppression (RPS) was introduced, showing success in improving the performance of the detection systems;
    
    \item The aforementioned contributions were merged into a single system, and validation of the resulting framework was performed across three different model scales over two real-world large-scale datasets. State-of-the-art performance for similar scale models was reported.
    
\end{itemize}

This paper is organized as follows: Section \ref{sec:relatedworks} presents a review of the related works on event representations, detectors, and data augmentation techniques. Then, in Section \ref{sec:methodology}, a discussion about the ideas proposed in this paper is performed. After that, the results are exposed in Section \ref{sec:results}. Finally, Section \ref{sec:Conclusion} summarizes the main achievements of this work and provides some insights about future works.

\section{Related Works}
\label{sec:relatedworks}

\subsection{Event Representations}
\label{sec:relatedworks:representations}




One of the most common and intuitive methods for event representation involves projecting events onto a 2D-pixel grid for modern CNN compatibility. An effective approach involves generating 2D grids based on timestamps \cite{sae},  \cite{hats}, \cite{inceptive}. Along the lines of this 2D concept, Event Frames — or Histograms — rely on event counts at each pixel location \cite{event_frame}, presenting in some cases channels separated by polarity \cite{eventCount1} or combinations of channels integrating polarity and timestamp features \cite{eventCount2}, \cite{cstr}.


Preserving temporal information from events often involves constructing dense representations segmented into distinct temporal windows, subsequently stacked to construct a 3D tensor. Voxel grids bin events across the time dimension, utilizing a bilinear kernel and interval normalization to weigh polarity contributions \cite{voxelgrid}. In contrast, Stacked Histograms replace this kernel by a simple event counting \cite{rvt}. Mixed-Density Event Stacks (MDES) offer a variation where bins encode different event densities within a single window segment \cite{mdes}. Hyper Histograms split temporal windows into smaller units, creating channels based on polarity and timestamp histograms \cite{aec}. Event Temporal Images map events within the 0 to 255 range to create grayscale images, incorporating distinct ranges to capture differing positive and negative event distributions \cite{dstdnet}.

Event Spike Tensor (EST) proposed an end-to-end learning process where MLPs are trained to find the best encoding according to a generalized 4D tensor, defined over the polarity and spatiotemporal domains \cite{est}. Asynchronous Attention Embedding employs an attention layer on events followed by a dilation-based 1D convolution used for data encoding \cite{astmnet}. EventPillars, inspired by PointPillars, is a trainable representation that treats events similarly to point clouds \cite{dmanet}. Event Representation through Gromov-Wasserstein Optimization (ERGO) employs Bayesian optimization over categorical variables, leveraging the Gromov-Wasserstein Discrepancy (GWD) as a key metric to assess the effectiveness of a particular event representation \cite{ergo12}.


Time-ordered Recent Events (TORE) volumes utilize First-In-First-Out (FIFO) queues with a depth of K, establishing a direct mapping to each pixel at every polarity. These volumes can be generated asynchronously, without a predefined time window \cite{tore}. Temporal Active Focus (TAF) aligns with TORE principles but integrates adaptive features for varying rates \cite{aed}.

Given the asynchronous nature of events, a proposed approach involves encoding them as nodes within graphs, with the connections between nodes defined as edges. This methodology allows efficient processing using Graph Neural Networks (GNNs) \cite{graph_event_first}. Voxel Cube introduced an alternative to event volumes where event accumulation within each micro-bin is binary, aiming to enhance temporal resolution specifically tailored for Spiking Neural Networks (SNNs) \cite{spikingDenseNet}. Group Tokens were specifically crafted for integration within Transformer-based architectures, involving the discretization of the event stream into intervals that are subsequently converted into patches. A 3x3 group convolution is then employed to embed the information into tokens effectively within this framework \cite{get}.

In this work, a memory-efficient and rapid event representation called VTEI is proposed to contribute to the design of an efficient and lightweight object detection framework. VTEI leverages a spatiotemporal volume to preserve temporal information, similar to Voxel Grids and Stacked Histograms. However, unlike these methods, VTEI represents each data unit within the volume using a limited number of values, similar to MDES. This approach results in a final representation characterized by high sparsity, low memory usage, low bandwidth, and low latency. Furthermore, VTEI effectively preserves sub-temporal dynamics within a given time window using minimal polarity information.

\subsection{Event-Based Object Detectors}
\label{sec:relatedworks:detectors}





One of the pioneering works on event-based object detection, Asynet, proposed leveraging the intrinsic spatial sparsity of event data by converting synchronous networks to asynchronous ones \cite{asynet}. Recently, a Graph Neural Network (GNN) approach called Asynchronous Event-Based GNN (AEGNN) was introduced, modeling events as spatio-temporal graphs with events as nodes and connections between neighboring events as edges. Processing is conducted through graph pooling and graph convolutions \cite{aegnn}. The potential of this approach has inspired the proposal of other similar networks \cite{sun2023event_gnn}, \cite{dagr}.

Spiking Neural Networks (SNNs), driven by spikes analogous to events, are acknowledged for their low power consumption, making them suitable for event-based camera applications. A hybrid SNN-ANN architecture was proposed, utilizing end-to-end training to leverage event information without intermediate representations \cite{hybrid-snn-ann}. The first SNN validated on real-world event data incorporated spiking variants of VGG \cite{vgg}, SqueezeNet \cite{squeezenet}, MobileNet \cite{mobilenet}, and DenseNet \cite{densenet} feature extractors attached to an SSD detection head, with DenseNet yielding the best performance \cite{spikingDenseNet}. By designing a full-spike residual block, the capability to directly train deep-SNNs for object detection improved, outperforming hybrid models and achieving real-time responses \cite{ems-yolo}. Building on spiking residual blocks, Spiking-Retinanet proposed an ANN-SNN detector \cite{spiking-retinanet}, while Spiking-YOLOv4 was developed using a CNN-to-SNN method \cite{spiking-yolov4}. Additionally, an SNN version of a Region Proposal Network (RPN) for object recognition was introduced \cite{attention-rpnsnn}. Spiking Fusion Object Detector (SFOD) was the first to adapt multi-scale feature fusion for SNNs using spiking data \cite{sfod}. Recently, a framework integrating the entire process from event sampling to feature extraction in an end-to-end fashion achieved competitive results with ANNs \cite{easnn}.

The Recurrent Event-camera Detector (RED) uses Squeeze-and-Excitation (SE) layers \cite{senet} for feature extraction and Convolutional Long Short-Term Memory (ConvLSTM) blocks for spatiotemporal data extraction, combined with an SSD detection head \cite{red_megapixel}. The Asynchronous Spatio-Temporal Memory Network (ASTMNet) comprises three components: Adaptive Temporal Sampling (ATS), Temporal Attention Convolutional Network (TACN), and Spatio-Temporal Memory. ATS samples events into bins based on an adaptive scheme related to the event frequency within an interval, while TACN aggregates events into an event representation called Asynchronous Attention Embedding. The Spatio-Temporal Memory module implements Recurrent-Convolutional (Rec-Conv) blocks following some convolutional layers \cite{astmnet}. The Agile Event Detector (AED) introduced a new event representation called Temporal Active Focus (TAF) to encode sparse event streams into dense tensors, enhancing temporal information extraction \cite{aed}. The Dual Memory Aggregation Network (DMANet) combines event information over different temporal ranges (short-term and long-term) with a learnable representation, EventPillars, for the detection task \cite{dmanet}. A YOLOv5 \cite{yolov5} detector was adapted to detect events encoded in a novel representation called Hyper Histograms, resulting in a remarkable reduction in terms of latency \cite{aec}.

Recurrent Vision Transformer (RVT) uses multi-axis attention \cite{maxvit} as a backbone, combined with ConvLSTMs \cite{convlstm} and YOLOX detection head for event-based detection \cite{rvt}. Enhancements to RVT through a self-labeling approach demonstrated further improvements \cite{leod}. A detector based on SWin-v2 \cite{swinv2} was proposed, utilizing event encodings optimized through the Gromov-Wasserstein Discrepancy. This approach achieved state-of-the-art performance without the need for recurrent modules \cite{ergo12}. HMNet proposed a multi-rate hierarchy with multiple cells to model local and global context information from objects with varying dynamics, introducing sparse cross-attention operations between features and events \cite{hmnet}. A transformer backbone featuring dual attention for spatial and polarity-temporal domains, paired with an event encoding focused on tokens, was also proposed \cite{get}. To address event sparsity, a mechanism for processing only tokens with meaningful information was recently introduced, including a version of the Self-Attention operation adjusted for unequal token sizes \cite{sast}. Recently, State Space Models (SSM) were introduced to replace RNN-cells for temporal modeling on detectors based on transformer backbones \cite{event-ssm}.

Most of the aforementioned works process event features through some network and then adopt detection heads used on frame data, where the YOLO family is the most common choice. From this family, YOLOv8 is a well-acknowledged Object Detector in terms of performance, real-time operation, and scalability \cite{yolov8}. However, it works only with frames, which, in turn, lacks resources for processing event-based data, such as temporal-based processing. As mentioned before, a common solution for this is to add recurrent cells to frame-based extractors, as done in \cite{red_megapixel}, \cite{astmnet}, and \cite{rvt}, for example. Then, in this work, an extension of the YOLOv8 framework is proposed to add compatibility with events processing and training. 

\subsection{Data Augmentation Techniques for Events}
\label{sec:relatedworks:dataaug}


EventDrop randomly drops events from an event stream, which can be applied to individual events, events within a specific spatial location, or events within a particular time window \cite{eventdrop}. Neuromorphic Data Augmentation (NDA) introduced an augmentation policy incorporating techniques such as Horizontal Flip, Rolling, Rotation, CutOut, and CutMix for training SNNs \cite{neuro_data_aug}. Spatio-temporal augmentation using random translation and time scaling was also proposed \cite{randomtimescale}. Temporal Event Shifting, which involves randomly reallocating events from a given frame to one of its prior frames, has proven beneficial for visual-aided force measurement \cite{force_event_aug}. Event Spatiotemporal Fragments combines the inversion of event fragments on spatiotemporal and polarity domains with spatiotemporal drift of some slices of events through a certain extent \cite{neural_fragments}. Moreover, a viewpoint transform based on translation and rotation, combined with spatiotemporal stretching to prevent information loss due to out-of-resolution events discarded during the initial transformation, was introduced for training SNNs \cite{viewpoint_trans_events}.

EventCopyDrop is an enhanced version of EventDrop. It includes an additional augmentation called EventCopy, which creates copies of events from one random region and places them in another random location within the stream \cite{aug_event_ssrl}. EventMix proposed an augmentation method based on mixing data from different event streams \cite{eventmix}. In RVT, Zoom-Out and Zoom-In augmentations were introduced to enhance event-based object detection \cite{rvt}. A framework combining geometric spatial augmentations with random temporal shifts and random polarity inversion was proposed \cite{cstr}.

Shadow Mosaic is a technique that simulates events with varying densities, referred to as Shadows, and arranges them into a Mosaic to create a larger sample \cite{aec}. ShapeAug introduces random occlusions to event data, enhancing the robustness of object recognition applications \cite{shapeaug}. Relevance Propagation Guidance (RPG) is employed to drop and mix events, resulting in the EventRPG augmentation method \cite{eventrpg}. EventAugment is an augmentation policy learning framework with 13 specific operations for event data, including flips, translations, crops, drops, and shear operations, targeting both spatial and temporal domains \cite{eventaugment}. Additionally, a temporal augmentation technique that involves dropping multiple sections of events within the temporal domain was proposed and evaluated for Lip-Reading applications \cite{temporal_mask_lipread}.

Despite numerous works presenting various approaches for event data augmentation across spatial, temporal, and polarity domains, certain phenomena associated with event-based camera operations that can affect the generalization of Deep Learning models remain underexplored. In real-world scenes, brightness distribution is often irregular, and polarity distribution can vary significantly from one scene to another. Additionally, types of noise discussed in \cite{original_dvs_paper} and the adjustable bias settings in pixel circuits \cite{sensitivity1} \cite{sensitivity2} can contribute to this variability. In this work, we propose a novel data augmentation model called Random Polarity Suppression to train Deep Learning models considering these variations.

\section{Methodology}
\label{sec:methodology}

This work proposes an event-based object detection framework based on YOLOv8. To support this, a novel event data encoding method is presented, aiming to convert event streams into CNN-suitable representations that can be calculated with low latency, resulting in tensors that require low bandwidth and memory. Moreover, a data augmentation technique involving the random suppression of positive and negative polarities is also introduced to enhance the system's performance.

\subsection{Volume of Ternary Event Images}
\label{proposed:vtei}

Event-based cameras function as 2D sensors capturing brightness variations at the pixel level. This process can be expressed mathematically as:

\begin{equation}
    \Delta L(x_{k}, y_{k}, t_{k}) \ge p_{k}C
    \label{eq:brightness_change}
\end{equation}

Here, $\Delta L$ denotes the logarithmic change in a photoreceptor's current (brightness) at pixel location $(x_{k}, y_{k})$ and time $t_{k}$. The polarity $p_{k} \in \{+1, -1\}$ indicates that a brightness change exceeding a threshold $C$ in absolute value triggers a positive or negative event \cite{original_dvs_paper}. An event is characterized by the tuple $e_{k} = (x_{k}, y_{k}, p_{k}, t_{k})$.

The sparse format of event streams poses a challenge for many current Deep Learning algorithms, requiring preprocessing for compatibility, as discussed in Section \ref{sec:relatedworks:representations}. With event-based cameras capable of operating at higher rates, driven by sensor resolution improvements \cite{are_1mp_required}, utilization of raw event data in downstream tasks can be complex. Efficient transformation involves sampling event streams at a consistent rate, partitioning them into sub-windows before conversion to dense tensors. This temporal binning strategy, effective in preserving temporal context from event streams, accommodates various conversion approaches such as applying a bilinear kernel to normalized timestamps \cite{voxelgrid}, \cite{red_megapixel}, event counting \cite{rvt}, or tracking the latest event at particular locations \cite{mdes}. In this study, leveraging the success of such methodologies, a variation of Event Volumes is adopted, focusing on computation time and memory requirements. The chosen encoding scheme focuses on sampling the last event at each spatiotemporal location, as done in MDES \cite{mdes}, but with uniform temporal bin sizes.

Initially, a tensor $I$ of dimensions $B$x$H$x$W$ is initialized with zeros, where $B$ represents the temporal bins and $H, W$ are spatial dimensions. For a stream $E = {{e_{k}}}_{k=0}^{N-1}$ containing $N$ events sampled at a consistent time window $T$, a temporal division into uniform intervals is executed using the formula:

\begin{equation}
    T_{k} = \floor{\frac{t_{k} - t_{0}}{t_{N} - t_{0}}}B
    \label{eq:bin}
\end{equation}

In this equation, $t_{0}$ represents the stream's initial timestamp, $t_{N}$ the final timestamp, and $T_{k}$ indicates the assigned temporal bin for the timestamp $t_{k}$. Subsequently, each pixel location $(X_{C}, Y_{C})$ within each temporal-based channel $i \in {0,1,..,B-1}$ is populated according to

\begin{equation}
    I(X_{C}, Y_{C}, i) = LastEvent(E_{X_{C}, Y_{C}, i})
    \label{eq:vtei}
\end{equation}

 where $E_{X_{C}, Y_{C}, i} \in E$ represents the event subset associated with spatiotemporal location $(X_{C}, Y_{C}, i)$, and $LastEvent$ extracts the polarity of the last event on it.

\begin{figure}[ht!]
    \centering
    \includegraphics[scale=0.6]{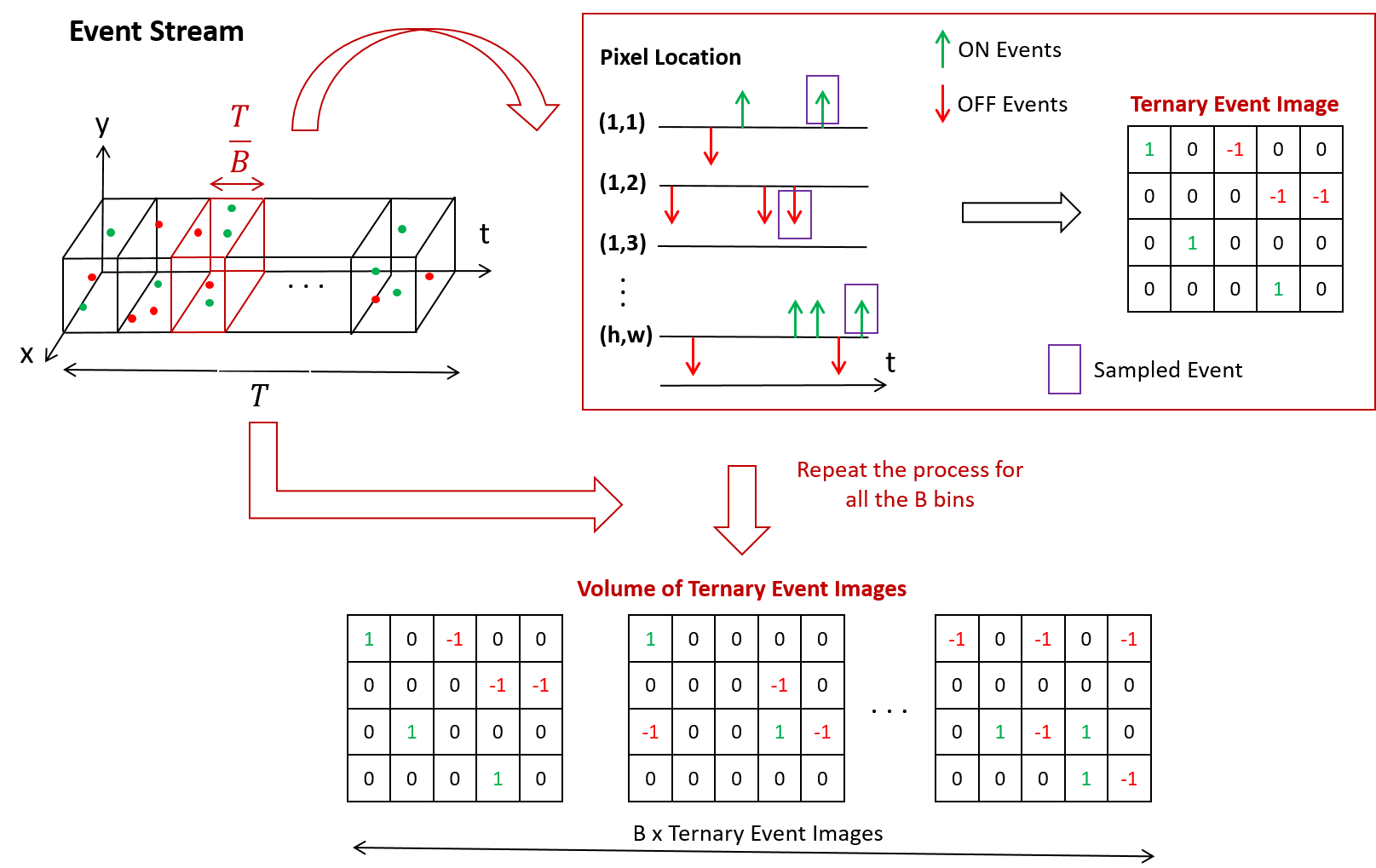}
    \caption{Working principle behind the Volume Ternary Event Image encoding.}
    \label{fig:vtei}
\end{figure}

Unlike MDES \cite{mdes}, which generates grayscale bins, this method encodes the last event data using two values, +1 and -1, retaining the original polarity without alteration. Designating the background as 0, each bin corresponds to a Ternary Event Image. Consequently, stacked bins form a Volume of Ternary Event Images (VTEI), as depicted in Figure \ref{fig:vtei}. Employing only three values benefits memory-constrained contexts like edge applications. Moreover, the background data can be discarded for compression purposes, minimizing bandwidth requirements for data transmission. A swift computation is ultimately anticipated, as the primary operation involves mapping each event to its relevant position in the dense grid.

\subsection{Recurrent YOLOv8 Architecture}
\label{proposed:ryolov8}

\begin{figure}[ht!]
    \centering
    \includegraphics[scale=0.32]{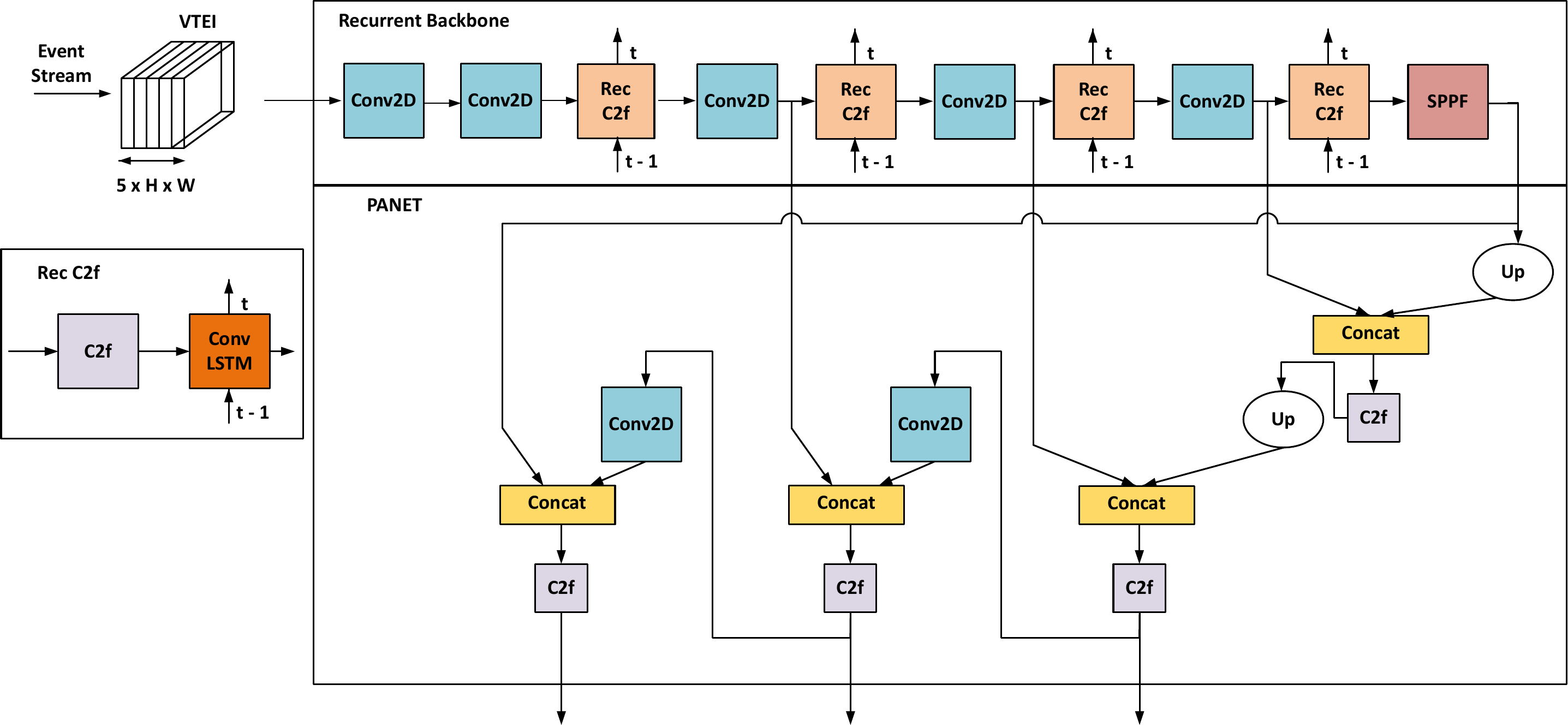}
    \caption{Overview of the Recurrent YOLOv8 architecture.}
    \label{fig:ryolov8}
\end{figure}

In Figure \ref{fig:ryolov8}, the general architecture of the Recurrent YOLOv8 is illustrated. Incoming event streams undergo conversion to VTEI tensors with 5 bins, following the method outlined in Section \ref{proposed:vtei}. These tensors feed the network's feature extractor, structured like the original YOLOv8 \cite{yolov8} but incorporating recurrent blocks and resizing certain convolutional blocks. The \textit{Conv2D} blocks function as standard convolutional layers for spatial feature downsampling. Starting from the 2nd stage, feature maps pass through \textit{Rec C2f} blocks for further refinement before downsampling. These blocks combine \textit{C2f} blocks, refining features in the channel domain, with a \textit{ConvLSTM} \cite{convlstm} block, which models long-range temporal relationships by incorporating current and past features. Subsequent to the final recurrent block, a Spatial Pyramid Pooling (SPP) block enriches features by combining multiple receptive fields \cite{spp}. The final features produced by this Recurrent Backbone are fed into YOLOv8's Path Aggregation Network (PANet) \cite{panet} for fusion and transmission to the 3-level detection heads.

\begin{figure}[ht!]
    \centering
    \includegraphics[scale=0.45]{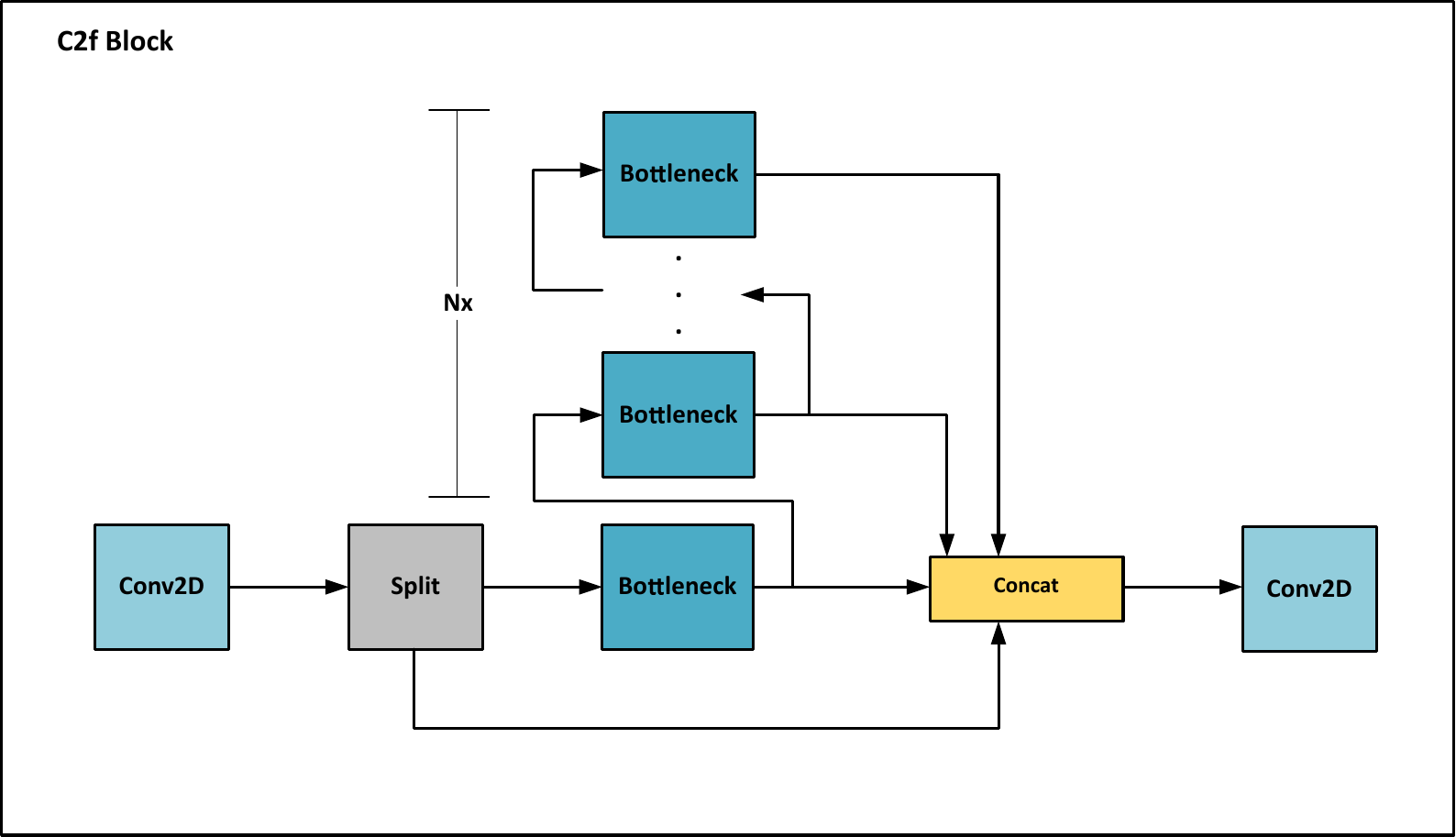}
    \caption{Structure of the CSP Bottleneck block with 2 convolutions from YOLOv8 \cite{yolov8}.}
    \label{fig:c2f}
\end{figure}

Figure \ref{fig:c2f} illustrates the architecture of a \textit{C2f} block \cite{yolov8}, which is an efficient version of a Cross-Stage Partial (CSP) Bottleneck block \cite{yolov4} with two convolutions. The initial convolution in this block adjusts the input channel count. Subsequently, a \textit{Split} block separates the feature into two groups with equal channels. One group undergoes processing through a sequence of \textit{N} Bottleneck blocks, with the same structure as the ResNet's blocks \cite{resnet}. Notably, the shortcut connections within these blocks are deactivated when incorporated into the PANET framework. Finally, the remaining split channels are merged with the output from each Bottleneck, followed by another convolution to reduce the number of channels.

The \textit{ConvLSTM} block depicted in Figure \ref{fig:ryolov8} implements a Long-Short Term Memory (LSTM) \cite{lstm} cell using 1x1 convolutions instead of the traditional Multi-Layer Perceptrons (MLP). This adaptation maintains the fundamental operational principle of an LSTM while enhancing its versatility to accommodate varying spatial dimensions. A similar structure was adopted on RVT \cite{rvt}. When considering an input $x$, along with the previous hidden $h_{t-1}$ and cell states $c_{t-1}$ of the block, the functionality of this cell can be expressed through the following equations:

\begin{align}
i = \sigma (Conv2D_{1x1}([x, h_{t-1}]))
    \label{eq:input_gate} \\
    r = \sigma (Conv2D_{1x1}([x, h_{t-1}]))
    \label{eq:remember_gate}\\
    o = \sigma (Conv2D_{1x1}([x, h_{t-1}]))
    \label{eq:output_gate}\\
    c = \tanh (Conv2D_{1x1}([x, h_{t-1}]))
    \label{eq:cell_gate}
\end{align}

In these equations, $i$, $r$, $o$, and $c$ represent the input, remember, output, and cell gates, respectively. Here, $Conv2D_{1x1}$ symbolizes a 2D convolution with a 1x1 kernel, $\sigma$ denotes a sigmoid activation function, and the square brackets indicate concatenation between two inputs. The current hidden and cell states, $h_{t}$ and $c_{t}$, respectively, can be determined by:

\begin{align}
    c_{t} = r*c_{t-1} + c*i
    \label{eq:cell_state1}\\
    h_{t} = o*\tanh(c_{t})
    \label{eq:cell_state2}
\end{align}

Additionally, $h_{t}$ is the output feature to be propagated to the subsequent block.

This work introduces three variants of the Recurrent YOLOv8 architecture displayed in Figure \ref{fig:ryolov8}: ReYOLOv8n, ReYOLOv8s, and ReYOLOV8m, representing nano, small, and medium scales, respectively. These scales align with the standards set by the original YOLOv8 \cite{yolov8}. Alongside incorporating recurrent cells, modifications are made regarding the channel count in each layer and the number of bottleneck blocks within each $C2f$, which are slightly adjusted compared to the reference models. Table \ref{tab:ryolov8s} provides a comprehensive overview of the modules within ReYOLOv8s, where $Cin$ and $Cout$ denote the number of input and output channels, respectively. ReYOLOv8n and ReYOLOv8m follow a similar framework, implementing distinct channel and width multipliers to rescale the number of channels and bottlenecks. In this sense, new models are created in a way similar to Compound Scaling \cite{efficientnet}. All $Conv2D$ operations are utilized for downsampling, utilizing a kernel size of 3 and a stride of 2.

\begingroup
\renewcommand{\arraystretch}{1.0}
\begin{table}[ht!]
\centering
\caption{Detailed architecture from ReYOLOv8s}
\label{tab:ryolov8s}
\begin{tabular}{|ccccccc|}
\hline
\multicolumn{1}{|c|}{\textbf{Node}} & \multicolumn{1}{c|}{\textbf{\begin{tabular}[c]{@{}c@{}}Input\\ Nodes\end{tabular}}} & \multicolumn{1}{c|}{\textbf{\begin{tabular}[c]{@{}c@{}}Block\\ Name\end{tabular}}} & \multicolumn{1}{c|}{\textbf{Cin}}        & \multicolumn{1}{c|}{\textbf{Cout}} & \multicolumn{1}{c|}{\textbf{\begin{tabular}[c]{@{}c@{}}Bottlenecks\\ (C2f only)\end{tabular}}} & \textbf{\begin{tabular}[c]{@{}c@{}}Residual\\ (C2f only)\end{tabular}} \\ \hline
\multicolumn{7}{|c|}{\textbf{Recurrent Backbone}}                                                                                                                                                                                                                                                                                                                                                                                                                        \\ \hline
\multicolumn{1}{|c|}{1}             & \multicolumn{1}{c|}{-}                                                              & \multicolumn{1}{c|}{Conv2D}                                                        & \multicolumn{1}{c|}{5}                   & \multicolumn{1}{c|}{24}            & \multicolumn{1}{c|}{-}                                                                         & -                                                                      \\ \hline
\multicolumn{1}{|c|}{2}             & \multicolumn{1}{c|}{1}                                                              & \multicolumn{1}{c|}{Conv2D}                                                        & \multicolumn{1}{c|}{24}                  & \multicolumn{1}{c|}{48}            & \multicolumn{1}{c|}{-}                                                                         & -                                                                      \\ \hline
\multicolumn{1}{|c|}{3}             & \multicolumn{1}{c|}{2}                                                              & \multicolumn{1}{c|}{C2f}                                                           & \multicolumn{1}{c|}{48}                  & \multicolumn{1}{c|}{48}            & \multicolumn{1}{c|}{2}                                                                         & -                                                                      \\ \hline
\multicolumn{1}{|c|}{4}             & \multicolumn{1}{c|}{3}                                                              & \multicolumn{1}{c|}{ConvLSTM}                                                      & \multicolumn{1}{c|}{48}                  & \multicolumn{1}{c|}{48}            & \multicolumn{1}{c|}{-}                                                                         & -                                                                      \\ \hline
\multicolumn{1}{|c|}{5}             & \multicolumn{1}{c|}{4}                                                              & \multicolumn{1}{c|}{Conv2D}                                                        & \multicolumn{1}{c|}{48}                  & \multicolumn{1}{c|}{88}            & \multicolumn{1}{c|}{-}                                                                         & -                                                                      \\ \hline
\multicolumn{1}{|c|}{6}             & \multicolumn{1}{c|}{5}                                                              & \multicolumn{1}{c|}{C2f}                                                           & \multicolumn{1}{c|}{88}                  & \multicolumn{1}{c|}{88}            & \multicolumn{1}{c|}{3}                                                                         & -                                                                      \\ \hline
\multicolumn{1}{|c|}{7}             & \multicolumn{1}{c|}{6}                                                              & \multicolumn{1}{c|}{ConvLSTM}                                                      & \multicolumn{1}{c|}{88}                  & \multicolumn{1}{c|}{88}            & \multicolumn{1}{c|}{-}                                                                         & -                                                                      \\ \hline
\multicolumn{1}{|c|}{8}             & \multicolumn{1}{c|}{7}                                                              & \multicolumn{1}{c|}{Conv2D}                                                        & \multicolumn{1}{c|}{88}                  & \multicolumn{1}{c|}{176}           & \multicolumn{1}{c|}{-}                                                                         & -                                                                      \\ \hline
\multicolumn{1}{|c|}{9}             & \multicolumn{1}{c|}{8}                                                              & \multicolumn{1}{c|}{C2f}                                                           & \multicolumn{1}{c|}{176}                 & \multicolumn{1}{c|}{176}           & \multicolumn{1}{c|}{3}                                                                         & -                                                                      \\ \hline
\multicolumn{1}{|c|}{10}            & \multicolumn{1}{c|}{9}                                                              & \multicolumn{1}{c|}{ConvLSTM}                                                      & \multicolumn{1}{c|}{176}                 & \multicolumn{1}{c|}{176}           & \multicolumn{1}{c|}{-}                                                                         & -                                                                      \\ \hline
\multicolumn{1}{|c|}{11}            & \multicolumn{1}{c|}{10}                                                             & \multicolumn{1}{c|}{Conv2D}                                                        & \multicolumn{1}{c|}{176}                 & \multicolumn{1}{c|}{344}           & \multicolumn{1}{c|}{-}                                                                         & -                                                                      \\ \hline
\multicolumn{1}{|c|}{12}            & \multicolumn{1}{c|}{11}                                                             & \multicolumn{1}{c|}{C2f}                                                           & \multicolumn{1}{c|}{344}                 & \multicolumn{1}{c|}{344}           & \multicolumn{1}{c|}{2}                                                                         & -                                                                      \\ \hline
\multicolumn{1}{|c|}{13}            & \multicolumn{1}{c|}{12}                                                             & \multicolumn{1}{c|}{ConvLSTM}                                                      & \multicolumn{1}{c|}{344}                 & \multicolumn{1}{c|}{344}           & \multicolumn{1}{c|}{-}                                                                         & -                                                                      \\ \hline
\multicolumn{1}{|c|}{14}            & \multicolumn{1}{c|}{13}                                                             & \multicolumn{1}{c|}{SPFF}                                                          & \multicolumn{1}{c|}{344}                 & \multicolumn{1}{c|}{344}           & \multicolumn{1}{c|}{-}                                                                         & -                                                                      \\ \hline
\multicolumn{7}{|c|}{\textbf{PANET}}                                                                                                                                                                                                                                                                                                                                                                                                                                     \\ \hline
\multicolumn{1}{|c|}{15}            & \multicolumn{1}{c|}{14}                                                             & \multicolumn{1}{c|}{Upsample}                                                      & \multicolumn{1}{c|}{344}                 & \multicolumn{1}{c|}{344}           & \multicolumn{1}{c|}{-}                                                                         & -                                                                      \\ \hline
\multicolumn{1}{|c|}{16}            & \multicolumn{1}{c|}{{[}15,10{]}}                                                    & \multicolumn{1}{c|}{Concat}                                                        & \multicolumn{1}{c|}{{[}344,176{]}}       & \multicolumn{1}{c|}{520}           & \multicolumn{1}{c|}{-}                                                                         & -                                                                      \\ \hline
\multicolumn{1}{|c|}{17}            & \multicolumn{1}{c|}{16}                                                             & \multicolumn{1}{c|}{C2f}                                                           & \multicolumn{1}{c|}{520}                 & \multicolumn{1}{c|}{176}           & \multicolumn{1}{c|}{2}                                                                         & False                                                                  \\ \hline
\multicolumn{1}{|c|}{18}            & \multicolumn{1}{c|}{17}                                                             & \multicolumn{1}{c|}{Upsample}                                                      & \multicolumn{1}{c|}{176}                 & \multicolumn{1}{c|}{176}           & \multicolumn{1}{c|}{-}                                                                         & -                                                                      \\ \hline
\multicolumn{1}{|c|}{19}            & \multicolumn{1}{c|}{{[}17,7{]}}                                                     & \multicolumn{1}{c|}{Concat}                                                        & \multicolumn{1}{c|}{{[}176,88{]}}        & \multicolumn{1}{c|}{264}           & \multicolumn{1}{c|}{-}                                                                         & -                                                                      \\ \hline
\multicolumn{1}{|c|}{20}            & \multicolumn{1}{c|}{19}                                                             & \multicolumn{1}{c|}{C2f}                                                           & \multicolumn{1}{c|}{264}                 & \multicolumn{1}{c|}{88}            & \multicolumn{1}{c|}{2}                                                                         & False                                                                  \\ \hline
\multicolumn{1}{|c|}{21}            & \multicolumn{1}{c|}{20}                                                             & \multicolumn{1}{c|}{Conv2D}                                                        & \multicolumn{1}{c|}{88}                  & \multicolumn{1}{c|}{88}            & \multicolumn{1}{c|}{-}                                                                         & -                                                                      \\ \hline
\multicolumn{1}{|c|}{22}            & \multicolumn{1}{c|}{{[}21,17{]}}                                                    & \multicolumn{1}{c|}{Concat}                                                        & \multicolumn{1}{c|}{{[}88,176{]}}        & \multicolumn{1}{c|}{264}           & \multicolumn{1}{c|}{-}                                                                         & -                                                                      \\ \hline
\multicolumn{1}{|c|}{23}            & \multicolumn{1}{c|}{22}                                                             & \multicolumn{1}{c|}{C2f}                                                           & \multicolumn{1}{c|}{264}                 & \multicolumn{1}{c|}{176}           & \multicolumn{1}{c|}{2}                                                                         & False                                                                  \\ \hline
\multicolumn{1}{|c|}{24}            & \multicolumn{1}{c|}{23}                                                             & \multicolumn{1}{c|}{Conv2D}                                                        & \multicolumn{1}{c|}{176}                 & \multicolumn{1}{c|}{176}           & \multicolumn{1}{c|}{}                                                                          &                                                                        \\ \hline
\multicolumn{1}{|c|}{25}            & \multicolumn{1}{c|}{{[}24,14{]}}                                                    & \multicolumn{1}{c|}{Concat}                                                        & \multicolumn{1}{c|}{{[}176,344{]}}       & \multicolumn{1}{c|}{520}           & \multicolumn{1}{c|}{}                                                                          &                                                                        \\ \hline
\multicolumn{1}{|c|}{26}            & \multicolumn{1}{c|}{25}                                                             & \multicolumn{1}{c|}{C2f}                                                           & \multicolumn{1}{c|}{520}                 & \multicolumn{1}{c|}{344}           & \multicolumn{1}{c|}{2}                                                                         & False                                                                  \\ \hline
\multicolumn{7}{|c|}{\textbf{Detection Head}}                                                                                                                                                                                                                                                                                                                                                                                                                            \\ \hline
\multicolumn{1}{|c|}{27}            & \multicolumn{1}{c|}{{[}20,23,26{]}}                                                 & \multicolumn{1}{c|}{Detect}                                                        & \multicolumn{1}{c|}{{[}88,176,344{]}} & \multicolumn{1}{c|}{-}              & \multicolumn{1}{c|}{-}                                                                         & -                                                                      \\ \hline
\end{tabular}
\end{table}

\endgroup

\subsection{Event Data Augmentation with Random Polarity Suppression}
\label{proposed:polaritySuppresion}

The polarity imbalance in event data can stem from various factors. First and foremost, changes in illumination are scene-dependent, making it difficult to ensure an equal distribution of positive and negative events corresponding to scene movements. Furthermore, electronic circuits within pixels are susceptible to noise across multiple stages, ranging from the photoreceptor sensor to the comparator and amplifier stages \cite{posch2014retinomorphic}. Even when other noise-related parameters are well-controlled, sporadic positive polarity noisy events have been documented \cite{original_dvs_paper}. Moreover, bias currents within different stages of an event-based camera pixel can be externally adjusted, potentially influencing sensitivity to distinct polarities in varying ways \cite{sensitivity1}, \cite{sensitivity2}. 


Taking this into consideration, a data augmentation technique that specifically targets the polarity domain is proposed. To train the detector effectively under unbalanced polarity scenarios, a probability $s$ will be introduced to suppress a specific polarity within each batch. Additionally, another probability $p$ will denote the likelihood of suppressing the positive polarities, with $(1 - p)$ representing the corresponding value for the negative ones. Considering the random variables $(r1, r2) \in [0,1]$, the subset of all pixels from the VTEI tensor $I$ with negative polarity $I_{n}$, and its positive counterpart $I_{p}$, the Random Polarity Suppression (RPS) technique will construct a new tensor $I^{'}$ based on the following condition:

\begin{equation}
    I^{'} = \begin{cases}
    I,& \text{if } r1 \geq s \text{, else}\\
    I_{p},& \text{if } r2\geq p  \text{, else }\\
    I_{n},&\text{otherwise}
\end{cases}
\end{equation}

\begin{figure}[ht!]
    \centering
    \includegraphics[scale=0.5]{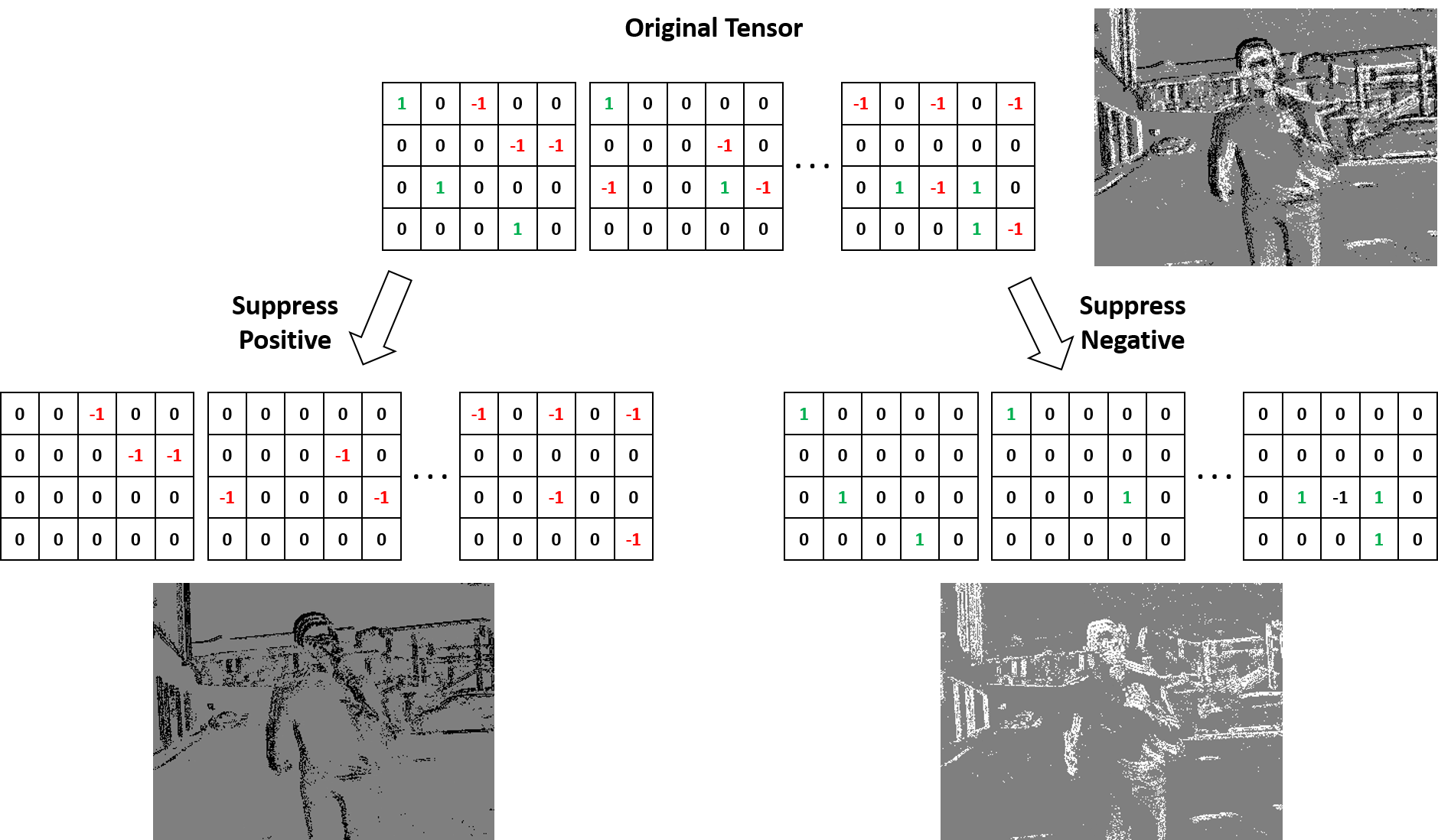}
    \caption{Example of Random Polarity Suppression's transformation on VTEI tensors. The grayscale corresponding images are also shown.}
    \label{fig:rps_example}
\end{figure}

For event encodings like VTEI, where there is a direct correlation between tensor values and their original event stream polarity, this augmentation can be applied post-conversion, reducing complexity compared to applying it to raw data. To ensure the consistency of the augmentation, the same modifications are applied across all temporal bins. Figure \ref{fig:rps_example} provides a schematic illustrating how RPS works on a given VTEI tensor, accompanied by grayscale representations of such images.

Given that this work utilizes VTEI encoding, the proposed augmentation technique reduces all information associated with a particular polarity. However, for encodings dependent on event counting, such as Stacked Histograms \cite{rvt}, it is feasible to implement this augmentation more gradually. This approach involves reducing the content of a specific polarity to a certain degree rather than entirely erasing it. To demonstrate the effectiveness and robustness of this technique, it is tested on real-world and extensive datasets known for their complexity in scenes and lighting conditions. Such datasets present a more challenging environment for evaluating the technique's performance and adaptability.

\subsection{Datasets}
\label{sec:datasets}

The object detection models in this work were validated using two substantial real-world event datasets, as outlined in Table \ref{tab:datasets}. The first dataset, PEDRo, is designed for person detection with a primary focus on Robotics applications. Recorded in Italy using a handheld camera, PEDRo captures individuals across diverse scenes, lighting conditions, and weather situations. The data was captured using a DAVIS346 camera with a resolution of 346x260 pixels. PEDRo is the sole real-world event-based large-scale dataset tailored specifically for Robotics applications to date \cite{pedro}. The second dataset, Prophesee's Generation 1 Automotive Dataset (GEN1), was recorded in France and encompasses various weather and illumination scenarios incorporating pedestrians and cars \cite{gen1}. While both datasets are significant in their respective applications and sizes, they exhibit complementary characteristics. GEN1 boasts a wider range of object classes; however, there exists an imbalance of approximately 5:1 between cars and pedestrians. Moreover, the pedestrian class is predominantly represented on smaller scales and towards the sides of the images, aligning with the expected viewpoint from a car's perspective. Conversely, PEDRo offers a more uniform representation of pedestrians across the pixel grid and showcases a greater diversity in terms of aspect ratios compared to GEN1 \cite{pedro}. Additionally, for GEN1, box filtering was applied to remove bounding boxes with one of the diagonals smaller than 30 or one of the sides smaller than 10, as done by \cite{red_megapixel}.

\begingroup
\renewcommand{\arraystretch}{1.1}
\begin{table}[ht!]
\centering
\caption{Datasets adopted in this work.}
\label{tab:datasets}
\begin{tabular}{|ccccc|}
\hline
\textbf{Dataset}                                                          & \textbf{Focus} & \textbf{Classes} & \textbf{Resolution} & \textbf{Labels} \\ \hline
\begin{tabular}[c]{@{}c@{}}PEDRO\\ \cite{pedro} \end{tabular} & Robotics       & 1                & 346x260             & 43k             \\ \hline
\begin{tabular}[c]{@{}c@{}}GEN1\\ \cite{gen1} \end{tabular}   & Automotive     & 2                & 304x240             & 255k            \\ \hline
\end{tabular}
\end{table}
\endgroup

\subsection{Training and Evaluation Procedure}
\label{subsec:train_and_eval}

To train the models described in Section \ref{sec:relatedworks:detectors}, Truncated Backpropagation Through Time (T-BPTT) \cite{bptt} was employed. During training, each dataset was segmented into clips with limited sequence lengths, and the memory cells were reset after each clip. During validation, complete original sequences were assessed, with memory cells being reset at the end of each sequence, aligning with methodologies observed in \cite{red_megapixel} and \cite{rvt}. Consistent application of data augmentation techniques was ensured across all frames within the same training sequence. The optimizer utilized was Stochastic Gradient Descent (SGD), with a momentum of 0.937 and linear learning rate decay.

In addition to T-BPTT, the training process closely adhered to the approach established within the YOLOv8 framework. A warm-up phase of 3 epochs was adopted to initiate training, consisting of a momentum of 0.8 and a bias learning rate of 0.1. The losses for box regression, classification, and Distribution Focal Loss (DFL) \cite{dflloss} maintained the same values as the original framework: 7.5, 0.5, and 1.5, respectively. Dataset-specific hyperparameters are detailed in Table \ref{tab:hyp}. All models were trained from scratch for 100 epochs. The sequence length is related to the T-BPTT setup. The image sizes from both datasets were adjusted to multiples of 32, aligning with the YOLOv8 anchors \cite{yolov8}. $LR0$ and $LRf$ reference the initial and final learning rates, respectively, determined by the prescribed schedule. Probability values for HFlip and Zoom-Out denote the likelihood of horizontal flip and zoom-out augmentations, with zoom-out scales randomly chosen between 1.0 and 1.2 for all cases. To speed up the training, validations were run every 10 epochs.

\begingroup
\renewcommand{\arraystretch}{1.1}
\begin{table}[ht!]
\centering

\caption{Hyperparameters adopted for the training of GEN1 and PEDRo datasets.}
\label{tab:hyp}
\begin{tabular}{|c|c|c|c|c|c|c|c|c|c|}
\hline
\textbf{Dataset}       & \textbf{Model} & \textbf{\begin{tabular}[c]{@{}c@{}}Sequence\\ Length\end{tabular}} & \textbf{\begin{tabular}[c]{@{}c@{}}Image\\  Size\end{tabular}} & \textbf{\begin{tabular}[c]{@{}c@{}}Batch\\  Size\end{tabular}} & \textbf{\begin{tabular}[c]{@{}c@{}}Weight \\ Decay\end{tabular}} & \textbf{LR0} & \textbf{LRf} & \textbf{HFlip} & \textbf{\begin{tabular}[c]{@{}c@{}}Zoom\\ Out\end{tabular}} \\ \hline
\multirow{3}{*}{GEN1}  & ReYOLOv8n       & 11                                                                 & 320x224                                                        & 48                                                             & 0.011                                                            & 0.01         & 0.0001       & 0.5            & 0.5                                                         \\ \cline{2-10} 
                       & ReYOLOv8s       & 11                                                                 & 320x224                                                        & 48                                                             & 0.011                                                            & 0.01         & 0.0001       & 0.5            & 0.5                                                         \\ \cline{2-10} 
                       & ReYOLOv8m       & 11                                                                 & 320x224                                                        & 36                                                             & 0.011                                                            & 0.01         & 0.0007       & 0.5            & 0.5                                                         \\ \hline
\multirow{3}{*}{PEDRO} & ReYOLOv8n       & 5                                                                  & 352x288                                                        & 48                                                             & 0.005                                                            & 0.07         & 0.0007       & 0.5            & 0.2                                                         \\ \cline{2-10} 
                       & ReYOLOv8s       & 5                                                                  & 352x288                                                        & 48                                                             & 0.005                                                            & 0.07         & 0.0007       & 0.5            & 0.2                                                         \\ \cline{2-10} 
                       & ReYOLOv8m       & 5                                                                  & 352x288                                                        & 48                                                             & 0.005                                                            & 0.07         & 0.0007       & 0.5            & 0.2                                                         \\ \hline
\end{tabular}
\end{table}
\endgroup

All training procedures were conducted utilizing a V100 GPU, while speed validations were executed on NVIDIA GTX1080ti and V100 GPUs for literature comparison. The entire development process was done using the PyTorch library \cite{pytorch}. The primary evaluation metric employed in this study was the Microsoft-Common Objects in Context (MS-COCO) mean Average Precision (mAP), specified as mAP@0.5:0.95. For this metric, a confidence threshold of 0.0001, consistent with YOLOv8, was utilized, along with a Non-Maximum Suppression (NMS) threshold of 0.7. During speed calculations, the confidence threshold was set to 0.25, and the Intersection over Union (IoU) was defined at 0.45. The final performance results were determined by evaluating the test subsets with the models that presented the highest mAP for the val subsets for each dataset.

\section{Results and Discussion}
\label{sec:results}

\subsection{Evaluation of VTEI}
\label{sec:format}



In order to assess the effectiveness of encoding event streams using VTEI, the test set from the GEN1 dataset was converted into three distinct formats commonly found in Event-Based Object Detection literature: Voxel Grids \cite{voxelgrid}, MDES \cite{mdes}, and Stacked Histograms (SHist) \cite{rvt}. The specific variation of Voxel Grids utilized by RED was employed \cite{red_megapixel}. The comparison excluded formats such as Hyper Histograms \cite{aec}, an extended version of SHist with additional channels; Event Temporal Images \cite{dstdnet}, similar to VTEI but with distinct accumulation and mapping processes; and asynchronous formats like TORE \cite{tore} and TAF tensors \cite{aed}, due to focusing on fixed time window encodings. Additionally, format-specific to different model categories like Voxel Cubes \cite{spikingDenseNet} for SNNs, Group Tokens \cite{get} for transformers, and graph-related representations \cite{graph_event_first}, \cite{aegnn}, \cite{dagr} were also excluded, as this study is centered on formats that can be adopted alongside convolutional architectures.

Table \ref{tab:data_summary} provides dimensions and the minimum byte counts for VTEI, SHist, and MDES for the GEN1 dataset. Compared to MDES and VTEI, SHist and Voxel Grids calculate polarities in separate channels before stacking them temporally, effectively doubling the number of channels. For GEN1, dimensions B, H, and W are 5, 240, and 304, respectively. The minimum byte count for each format was computed based on the Coordinate List (COO) compression, where each non-zero is encoded by its coordinate and data content. For the GEN1, spatial dimensions are encoded in 17 bits. The data content for VTEI and MDES can be represented in binary format, while the number of bins can be encoded in 3 bits. Then, summing it up, VTEI and MDES require 3 bytes per non-zero entry. SHist needs an additional bit to represent the channel domain, while the data itself is encoded in 8 bits, totaling 4 bytes per entry. Voxel Grids is similar to SHist but utilizes a sum function over normalized timestamps instead of event counting. Hence, with a decimal range requirement, a half-precision floating-point format (16 bits) is required, which is the smallest present in libraries like Pytorch, adding an extra byte for data encoding. Overall, the merging of polarities into the same channels and employing a narrower data value range positioned VTEI and MDES as superior options in terms of memory requirements.

\begingroup
\renewcommand{\arraystretch}{1.1}
\begin{table}[ht!]
\centering
\caption{Comparison between different event encodings in dimensions and minimum number of bytes required to encode data from the GEN1 dataset. The calculations consider B=5, H=240, and W=304.}
\label{tab:data_summary}
\begin{tabular}{|ccc|}
\hline
\textbf{Data Format}                                      & \textbf{Dimensions} & \textbf{Minimum \# Bytes} \\ \hline
\hline
\textbf{VTEI} (\textbf{this work})    & BxHxW               & \textbf{3}                                                         \\ \hline
SHist (\cite{rvt})       & 2BxHxW              & 4                                                                  \\ \hline
MDES (\cite{mdes})        & BxHxW               & \textbf{3}                                                         \\ \hline
\begin{tabular}[c]{@{}c@{}}Voxel Grid\\ (\cite{voxelgrid}, \cite{red_megapixel}) \end{tabular} & 2BxHxW              & 5                                                                  \\ \hline
\end{tabular}
\end{table}
\endgroup


To better highlight the capabilities of each format, a further analysis was performed, taking into consideration the recording from the GEN1's test set with the biggest number of events. All analyses were conducted on 50ms samples extracted from the original recordings. The outcomes, summarized in Table \ref{tab:data_performance}, are categorized into three sections detailing the average, maximum, and minimum event counts across all chunks. The performance data were gathered using an Intel Xeon Gold 6230R CPU on an Ubuntu 20.04.6 LTS operating system with 251GB of RAM. Regarding latency for a 50ms window (Latency@50ms), based on the value most commonly adopted in the literature, VTEI emerged as the top performer compared to other formats in all scenarios, with the disparity increasing as the number of events increased. Notably, VTEI was 1.54x faster on average than SHist, rising to 2.17x at the maximum event count case. Comparatively, VTEI showcased an average 2.24x speed-up over MDES, scaling to 2.5x at the peak event number, while these metrics were 3.4x and 4.0x compared to Voxel Grids, respectively. The Event Rate in Mev/s, reflecting the event processing capabilities, also favored VTEI across all scenarios, with an average rate approximately 2.39x higher than alternative formats, with a maximum of 2.89x. The event rate increased 3.23x from the minimum to the maximum event counts, highlighting the scalability of VTEI.

This table also reveals the number of non-zero elements after encoding for the three scenarios. The "Encoded Size" field represents the disk space required to encode each scenario based on the number of bytes shown in Table \ref{tab:data_summary}. The Compression Ratio is derived by dividing the disk size of the reference events in each section by their respective Encoded Sizes. It can be seen that VTEI achieves a compression ratio ranging from 2.53x to 2.95x, which is, on average, 1.79x higher than SHist, 2.23x better than Voxel Grids, and approximately 92\% as efficient as MDES. It is worth mentioning that MDES adopts non-uniform temporal bins, whose lengths decrease by powers of two, where some intervals are larger than those seen with VTEI. By sampling only the latest events in each bin, more events tend to be ignored at larger intervals. The larger sub-interval of MDES comprises half of the whole time window, contributing to a smaller number of non-zero elements and related metrics. Additionally, to provide communication-related insights, a Bandwidth (BW) field is included, calculated by dividing the Encoded Sizes by the sum of the total sampling period (assumed to be 50ms here) and the latencies from the conversion process. In this context, VTEI’s bandwidth ranges from 1.14MB/s to 9.54MB/s, which is, on average, 1.16xlower than SHist, 2.1x better than Voxel Grid, and only 1.13x higher than MDES. In conclusion, VTEI stands out as the best choice for latency and event processing rate among the options presented, requiring the lowest number of bytes for data encoding (alongside MDES), and being competitive with MDES regarding memory requirements.


\begingroup
\renewcommand{\arraystretch}{1.1}
\begin{table}[ht!]
\centering
\caption{Comparison between different event encodings based on chunks of event streams with 50ms from the test set of the GEN1 dataset in terms of speed, compression rate, and bandwidth.}
\label{tab:data_performance}
\begin{tabular}{|ccccccc|}
\hline
\multicolumn{7}{|c|}{\textbf{Average \#Events Processed = 192,063 (0.73MB)}}                                                                                                                                                                                                                                                                                                                                                                                                                                                                                              \\ \hline
\multicolumn{1}{|c|}{\textbf{\begin{tabular}[c]{@{}c@{}}Data\\ Format\end{tabular}}} & \multicolumn{1}{c|}{\textbf{\begin{tabular}[c]{@{}c@{}}Latency\\ @50ms\end{tabular}}} & \multicolumn{1}{c|}{\textbf{\begin{tabular}[c]{@{}c@{}}Event\\ Rate (Mev/s)\end{tabular}}} & \multicolumn{1}{c|}{\textbf{Non-zeros}} & \multicolumn{1}{c|}{\textbf{\begin{tabular}[c]{@{}c@{}}Encoded\\ Size(MB)\end{tabular}}} & \multicolumn{1}{c|}{\textbf{\begin{tabular}[c]{@{}c@{}}Compression\\ Ratio\end{tabular}}} & \textbf{\begin{tabular}[c]{@{}c@{}}BW\\ (MB/s)\end{tabular}} \\ \hline
\multicolumn{1}{|c|}{VTEI}                                                           & \multicolumn{1}{c|}{\textbf{1.5ms}}                                                   & \multicolumn{1}{c|}{\textbf{128.04}}                                                       & \multicolumn{1}{c|}{96,017}             & \multicolumn{1}{c|}{0.27}                                                                & \multicolumn{1}{c|}{2.67}                                                                 & 5.33                                                         \\ \hline
\multicolumn{1}{|c|}{SHist}                                                          & \multicolumn{1}{c|}{2.3ms}                                                            & \multicolumn{1}{c|}{83.14}                                                                 & \multicolumn{1}{c|}{125,234}            & \multicolumn{1}{c|}{0.48}                                                                & \multicolumn{1}{c|}{1.53}                                                                 & 9.13                                                         \\ \hline
\multicolumn{1}{|c|}{MDES}                                                           & \multicolumn{1}{c|}{3.4ms}                                                            & \multicolumn{1}{c|}{57.61}                                                                 & \multicolumn{1}{c|}{\textbf{89,532}}    & \multicolumn{1}{c|}{\textbf{0.26}}                                                       & \multicolumn{1}{c|}{\textbf{2.86}}                                                        & \textbf{4.80}                                                \\ \hline
\multicolumn{1}{|c|}{Voxel Grid}                                                     & \multicolumn{1}{c|}{5.1ms}                                                            & \multicolumn{1}{c|}{37.66}                                                                 & \multicolumn{1}{c|}{125,232}            & \multicolumn{1}{c|}{0.60}                                                                & \multicolumn{1}{c|}{1.23}                                                                 & 10.84                                                       \\ \hline
\multicolumn{7}{|c|}{\textbf{Maximum \#Events Processed = 327,030 (1.25MB)}}                                                                                                                                                                                                                                                                                                                                                                                                                                                                                              \\ \hline
\multicolumn{1}{|c|}{\textbf{\begin{tabular}[c]{@{}c@{}}Data\\ Format\end{tabular}}} & \multicolumn{1}{c|}{\textbf{\begin{tabular}[c]{@{}c@{}}Latency\\ @50ms\end{tabular}}} & \multicolumn{1}{c|}{\textbf{\begin{tabular}[c]{@{}c@{}}Event\\ Rate (Mev/s)\end{tabular}}} & \multicolumn{1}{c|}{\textbf{Non-zeros}} & \multicolumn{1}{c|}{\textbf{\begin{tabular}[c]{@{}c@{}}Encoded\\ Size(MB)\end{tabular}}} & \multicolumn{1}{c|}{\textbf{\begin{tabular}[c]{@{}c@{}}Compression\\ Ratio\end{tabular}}} & \textbf{\begin{tabular}[c]{@{}c@{}}BW\\ (MB/s)\end{tabular}} \\ \hline
\multicolumn{1}{|c|}{VTEI}                                                           & \multicolumn{1}{c|}{\textbf{1.8ms}}                                                   & \multicolumn{1}{c|}{\textbf{181.68}}                                                       & \multicolumn{1}{c|}{172,654}            & \multicolumn{1}{c|}{0.49}                                                                & \multicolumn{1}{c|}{2.53}                                                                 & 9.54                                                         \\ \hline
\multicolumn{1}{|c|}{SHist}                                                          & \multicolumn{1}{c|}{3.9ms}                                                            & \multicolumn{1}{c|}{83.85}                                                                 & \multicolumn{1}{c|}{219,848}            & \multicolumn{1}{c|}{0.84}                                                                & \multicolumn{1}{c|}{1.49}                                                                 & 15.56                                                        \\ \hline
\multicolumn{1}{|c|}{MDES}                                                           & \multicolumn{1}{c|}{4.5ms}                                                            & \multicolumn{1}{c|}{72.67}                                                                 & \multicolumn{1}{c|}{\textbf{145,305}}   & \multicolumn{1}{c|}{\textbf{0.42}}                                                       & \multicolumn{1}{c|}{\textbf{3.00}}                                                        & \textbf{7.63}                                                \\ \hline
\multicolumn{1}{|c|}{Voxel Grid}                                                     & \multicolumn{1}{c|}{7.2ms}                                                            & \multicolumn{1}{c|}{45.42}                                                                 & \multicolumn{1}{c|}{219,846}            & \multicolumn{1}{c|}{1.05}                                                                & \multicolumn{1}{c|}{1.19}                                                                 & 18.33                                                        \\ \hline
\multicolumn{7}{|c|}{\textbf{Minimum \#Events Processed = 44,962 (0.17MB)}}                                                                                                                                                                                                                                                                                                                                                                                                                                                                                               \\ \hline
\multicolumn{1}{|c|}{\textbf{\begin{tabular}[c]{@{}c@{}}Data\\ Format\end{tabular}}} & \multicolumn{1}{c|}{\textbf{\begin{tabular}[c]{@{}c@{}}Latency\\ @50ms\end{tabular}}} & \multicolumn{1}{c|}{\textbf{\begin{tabular}[c]{@{}c@{}}Event\\ Rate (Mev/s)\end{tabular}}} & \multicolumn{1}{c|}{\textbf{Non-zeros}} & \multicolumn{1}{c|}{\textbf{\begin{tabular}[c]{@{}c@{}}Encoded\\ Size(MB)\end{tabular}}} & \multicolumn{1}{c|}{\textbf{\begin{tabular}[c]{@{}c@{}}Compression\\ Ratio\end{tabular}}} & \textbf{\begin{tabular}[c]{@{}c@{}}BW\\ (MB/s)\end{tabular}} \\ \hline
\multicolumn{1}{|c|}{VTEI}                                                           & \multicolumn{1}{c|}{\textbf{0.8ms}}                                                   & \multicolumn{1}{c|}{\textbf{56.20}}                                                        & \multicolumn{1}{c|}{20,310}             & \multicolumn{1}{c|}{0.058}                                                               & \multicolumn{1}{c|}{2.95}                                                                 & 1.14                                                         \\ \hline
\multicolumn{1}{|c|}{SHist}                                                          & \multicolumn{1}{c|}{1.0ms}                                                            & \multicolumn{1}{c|}{44.96}                                                                 & \multicolumn{1}{c|}{29,268}             & \multicolumn{1}{c|}{0.11}                                                                & \multicolumn{1}{c|}{1.54}                                                                 & 2.19                                                         \\ \hline
\multicolumn{1}{|c|}{MDES}                                                           & \multicolumn{1}{c|}{1.8ms}                                                            & \multicolumn{1}{c|}{24.98}                                                                 & \multicolumn{1}{c|}{\textbf{19,832}}    & \multicolumn{1}{c|}{\textbf{0.057}}                                                      & \multicolumn{1}{c|}{\textbf{3.02}}                                                        & \textbf{1.10}                                                \\ \hline
\multicolumn{1}{|c|}{Voxel Grid}                                                     & \multicolumn{1}{c|}{2.1ms}                                                            & \multicolumn{1}{c|}{21.41}                                                                 & \multicolumn{1}{c|}{29,266}             & \multicolumn{1}{c|}{0.14}                                                                & \multicolumn{1}{c|}{1.23}                                                                 & 2.68                                                         \\ \hline
\end{tabular}
\end{table}
\endgroup

\subsection{Evaluation of Random Polarity Suppression}
\label{sec:aug}

\begin{figure}[ht!]
    \centering
    \includegraphics[scale=0.3]{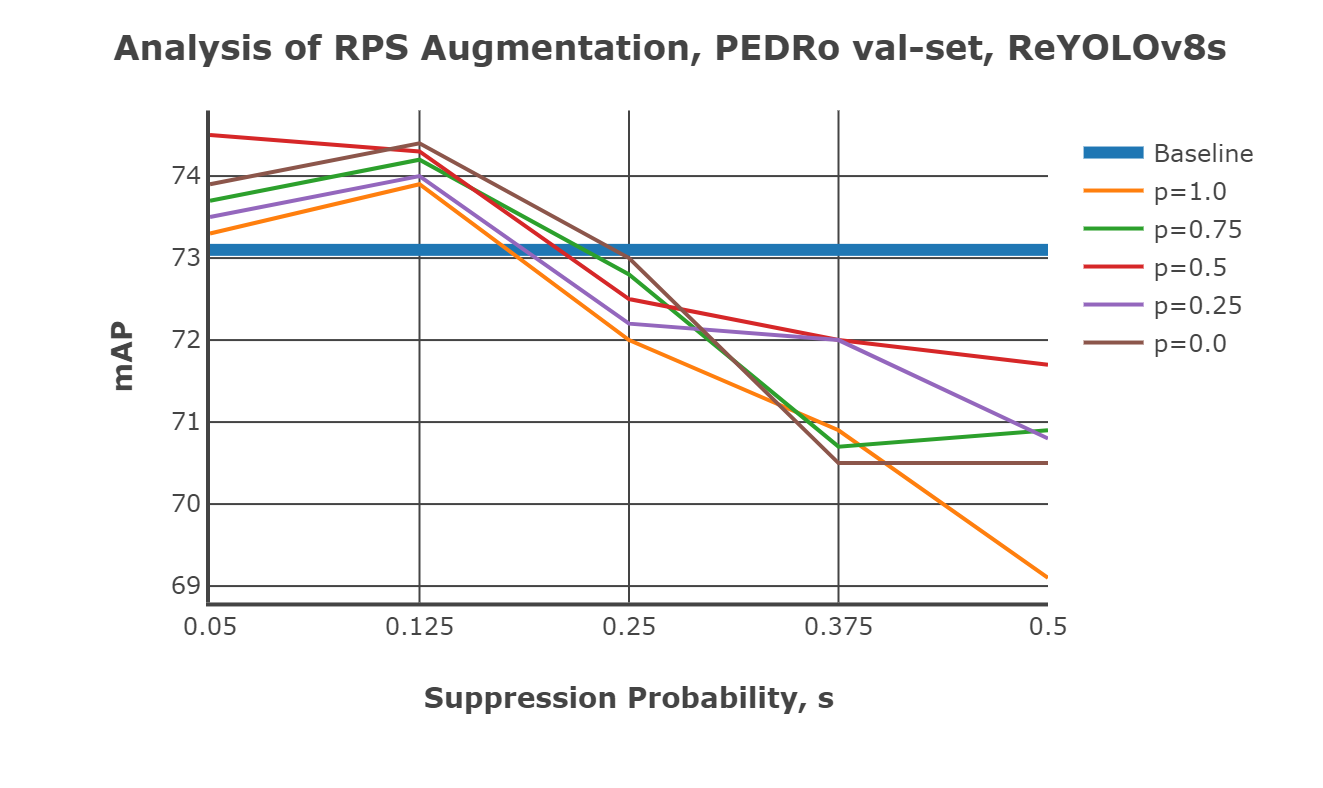}
    \caption{Comparison of the mAP from ReYOLOv8s by sweeping the suppression polarity, $s$,  given fixed probabilities of suppressing the positive polarity rather than the negative ones, given by $p$, for the PEDRO dataset's validation set.}
    \label{fig:pedro_rps_val}
\end{figure}

The impact of applying Polarity Suppression (RPS) on models was evaluated by comparing the baseline mean Average Precision (mAP) of ReYOLOv8s with the results using RPS. The positive polarity suppression probability $p$ was initially set at five levels: {0.0, 0.25, 0.50, 0.75, 1.0}. Different suppression probabilities $s$ from {0.05, 0.125, 0.25, 0.375, 0.5} were tested for each $p$ value. Five runs were conducted for both the baseline and each RPS combination, and average results were reported to account for variability. Figure \ref{fig:pedro_rps_val} displays the validation set results from the PEDRo dataset. The findings show that improvements generally occur up to a 12.5\% suppression probability, suggesting that RPS enhances model performance when used as a small disturbance. Larger suppression levels may degrade the quality of training samples and negatively impact performance. The most significant enhancements across all $p$ values were seen at $s=0.05$ and $s=0.125$. Notably, on average, $p=0.50$ led to the greatest improvements, followed by $p=0.0$, $p=0.75$, $p=0.25$, and $p=1.0$. There is no clear pattern regarding which polarity should be suppressed, as similar performance levels were observed for $p=0.0$, which represents a total negative polarity suppression, and $p=0.75$, an aggressive positive polarity suppression. A comparable trend was seen for $p=0.25$ and $p=1.0$, which involve opposite types of data manipulation but resulted in similar outcomes at $s=0.125$. These results suggest that objects in the PEDRo dataset have a diverse polarity distribution and are not particularly biased towards one polarity. This also explains why the most balanced scenario, $p=0.50$, led to the best results on average.

\begin{figure}[ht!]
    \centering
    \includegraphics[scale=0.3]{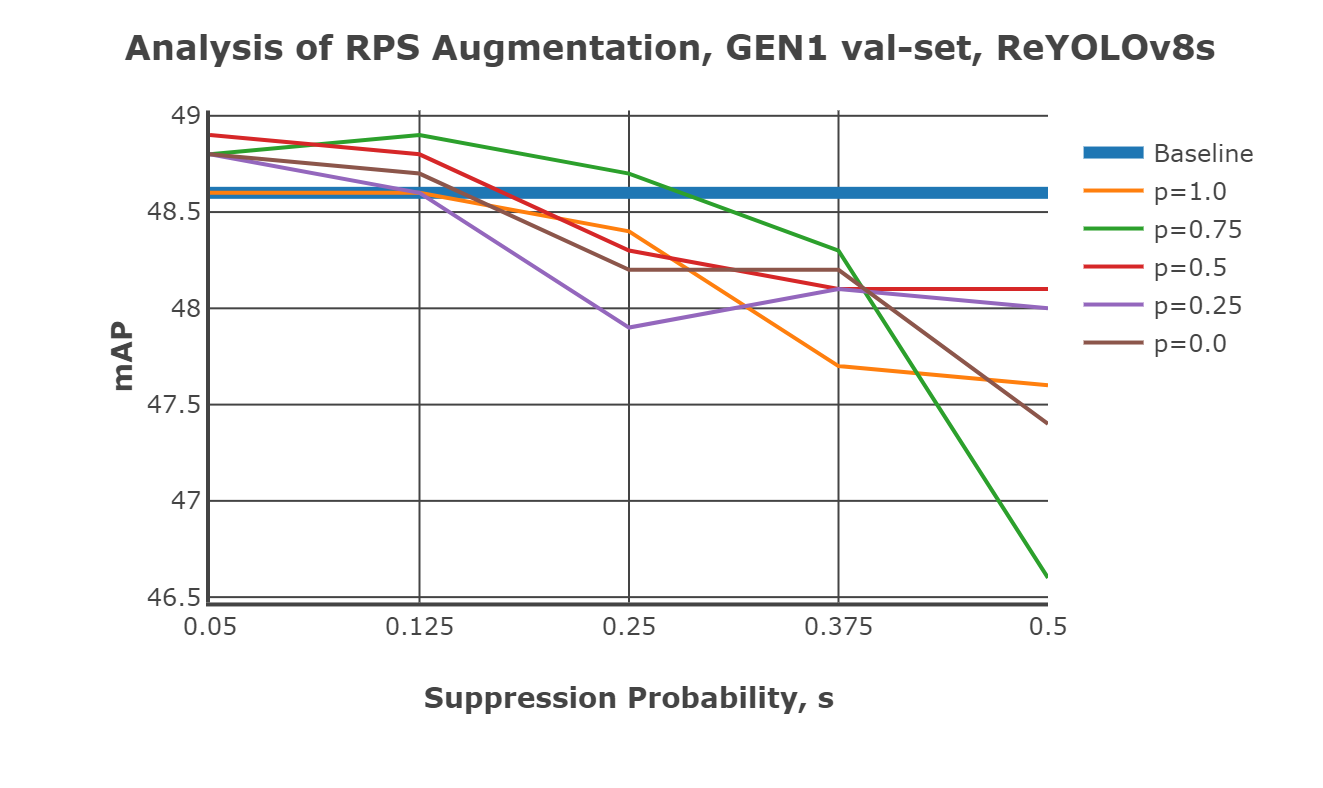}
    \caption{Comparison of the mAP from ReYOLOv8s by sweeping the suppression polarity, $s$,  given fixed probabilities of suppressing the positive polarity rather than the negative ones, given by $p$, for the GEN1 dataset's validation set.}
    \label{fig:gen1_rps_val}
\end{figure}

Figure \ref{fig:gen1_rps_val} shows the results of applying RPS to the GEN1 validation set, using the same procedure as for PEDRo. A similar trend to Figure \ref{fig:pedro_rps_val} is observed, where improvements in mean Average Precision (mAP) are more significant for suppression probabilities below 12.5\%. Except for $p=0.75$, no improvements were observed for $s=0.25$. When examining positive and negative suppression separately, the balanced scenario with $p=0.50$ performed better than most other values, similar to the findings for PEDRo. However, an exception was observed for $p=0.75$, which achieved the highest average improvement and a peak enhancement comparable to $p=0.50$. At $p=0.0$ and $p=0.25$, there was a noticeable increase in mAP at $s=0.05$, with steep decreases after $s=0.125$. No improvement was observed for $p=1.0$. 


Table \ref{tab:final_rps_table} expands on the previous analysis to include other scales of ReYOLOv8, detailing the combinations of suppression and positive/negative probabilities that resulted in the most significant improvements for each case. Due to time constraints, single runs were conducted for this analysis. Since more runs were performed for the ReYOLOv8s models, the baseline and RPS cases with the highest mAPs on each validation set were selected. When individually assessing the models, it was observed that GEN1 showed comparable improvement across all scales. In contrast, for PEDRo, the Nano and Medium models exhibited more substantial improvements compared to the Small variant. Interestingly, a common trend across all models pointed towards greater enhancements at lower suppression probabilities, consistent with the patterns identified in Figures \ref{fig:pedro_rps_val} and \ref{fig:gen1_rps_val}. Moreover, a balanced probability of positive and negative suppression was usually the most effective. Variations in scenes due to changes in illumination or camera instability, which contribute to polarity imbalance, along with other factors highlighted in Section \ref{sec:aug}, can impact the efficacy of the suppression technique on object detection performance. The stochastic nature of data augmentation, where samples are randomly subjected to transformations like RPS, further complicates predicting whether suppressing positive or negative polarities will yield improvements. After suppression, the modified tensors may not align with the intended polarity distribution. In conclusion, the analysis suggests that maintaining a sufficiently low level of general suppression may lead to enhancements at different positive and negative suppression levels, with a balanced probability being the safest choice for the latter.

\begingroup
\renewcommand{\arraystretch}{1.1}

\begin{table}[ht!]
\centering
\caption{Analysis of the effect of Random Polarity Suppression over the mAP of the models presented in this work.}
\label{tab:final_rps_table}
\begin{tabular}{|ccccccc|}
\hline
\textbf{Dataset}                & \textbf{Model} & \textbf{Suppression} & \textbf{Positive} & \textbf{\begin{tabular}[c]{@{}c@{}}Baseline \\ mAP\end{tabular}} & \textbf{\begin{tabular}[c]{@{}c@{}}mAP \\ with RPS\end{tabular}} & \textbf{Improvement} \\ \hline
\multirow{3}{*}{\textbf{GEN1}}  & ReYOLOv8n       & 0.25                 & 0.25              & 45.9                                                             & 46.3                                                             & +0.4                 \\
                                & ReYOLOv8s       & 0.05                & 0.50              & 48.1                                                             & 48.3                                                             & +0.2                 \\
                                & ReYOLOv8m       & 0.05                 & 0.50            & 49.0                                                             & 49.4                                                             & +0.4                 \\ \hline
\multirow{3}{*}{\textbf{PEDRo}} & ReYOLOv8n       & 0.125                 & 0.25             & 59.0                                                            & 63.9                                                             & +4.9                 \\
                                & ReYOLOv8s       & 0.05                 & 0.50             &  64.5                                                          & 65.5                                                             &  +1.0                \\
                                & ReYOLOv8m       & 0.125                 & 0.50             &  66.5                                                            & 69.1                                                             & +2.6                 \\ \hline
\end{tabular}
\end{table}

\endgroup


\subsection{Comparison with the state-of-the-art}
\label{sec:literature}

Table \ref{tab:pedro_literature} compares the three models introduced in this study, namely ReYOLOv8n, ReYOLOv8s, and ReYOLOv8m, with the state-of-the-art YOLOv8x-based model for PEDRo. All models exhibited improvements in mean Average Precision (mAP), ranging from 9\% to 18\%, requiring significantly fewer parameters - 14.5x and 3.8x, respectively. This notable performance improvement at a lower number of parameters can be attributed to the integration of long-range temporal modeling in the models of this study, a feature lacking in the benchmark YOLOv8x-based model. Given that the original work did not include runtime information \cite{pedro}, the inference times for YOLOv8x processing the same VTEI tensors as ReYOLOv8 were reported here. From this comparative analysis, the models in this study demonstrated an average speed-up of 1.46x, highlighting their efficiency. Furthermore, as shown in Table \ref{tab:final_rps_table}, the ReYOLOv8n implementation without RPS achieved a similar mAP to YOLOv8x, differing by only 0.6\%. However, after applying RPS, the gap between the models widened to 9\%. This indicates that data augmentation through RPS significantly contributed to the improvements observed in this study, in addition to utilizing memory cells.

\begingroup
\renewcommand{\arraystretch}{1.1}

\begin{table}[ht!]
\centering
\caption{Comparison with the literature for PEDRO dataset.}
\label{tab:pedro_literature}
\begin{tabular}{|ccccc|}
\hline
\textbf{Model}                                                              & \textbf{Network} & \textbf{Parameters} & \textbf{mAP} & \textbf{Runtime} \\ \hline
\textbf{\begin{tabular}[c]{@{}c@{}}ReYOLOv8n\\ (this work)\end{tabular}}     & CNN + RNN        & \textbf{4.7M}                &  63.9         & \textbf{9.2ms}           \\
\textbf{\begin{tabular}[c]{@{}c@{}}ReYOLOv8s\\ (this work)\end{tabular}}     & CNN + RNN        & 8.4M                &  65.5       & 10.4ms           \\
\textbf{\begin{tabular}[c]{@{}c@{}}ReYOLOv8m\\ (this work)\end{tabular}}     & CNN + RNN        & 18.1M                 &  \textbf{69.1}         & 12.3ms          \\
\begin{tabular}[c]{@{}c@{}}YOLOv8x \\ \cite{pedro} \end{tabular} & CNN              & 68.2M               & 58.6        &  17.6ms                \\ \hline
\end{tabular}
\end{table}
\endgroup

Table \ref{tab:gen1_table} presents a comparative analysis of the state-of-the-art models for the GEN1 dataset. Models are classified into Nano, Small, and Medium scales based on their trainable parameter ranges (5M, 15M, and 45M, respectively), a common practice in Computer Vision literature. Inference times are reported separately for NVIDIA's V100 GPU and a group consisting of closely related devices: NVIDIA's GTX1080ti, Titan XP, and GTX980. Only models with mAP exceeding 40.0 are considered, and the best results are highlighted. Firstly, analyzing the Nano models, ReYOLOv8n achieves a 5\% improvement in mAP compared to RVT-T, requiring only an additional 0.3M parameters while reducing latency by 0.20ms. When comparing the Small models, ReYOLOv8s also demonstrates superior performance, with a 2.8\% increase in mAP compared to HMNet-L1, demanding around 27\% fewer parameters but being 2.4x slower. On the medium scale, ReYOLOv8m outperforms SAST-CB \cite{sast} by 2.5\%, but also with fewer parameters. The superior performance of ReYOLOv8 can be attributed to its foundation on the YOLOv8 baseline, known to outperform other detectors like YOLOX \cite{yolox}, which serves as the detection head for RVT \cite{rvt}, GET \cite{get}, and SAST-CB \cite{sast} models. Models exceeding 45M parameters were omitted, as they did not yield substantial gains compared to the Medium-scaled models. Notable exceptions include DSTDNet-X \cite{dstdnet}, which has 100M parameters and has a mAP similar to ReYOLOv8m, and ERGO12 \cite{ergo12}, a model based on a pre-trained SWinv2 \cite{swinv2} transformer that achieves a mAP of 50.4 with 59.6M parameters and a latency of 77ms.

In terms of runtime, it can be seen that the models are competitive when compared to other approaches that deploy RNNs, where ReYOLOv8n is 2\% faster than RVT-T, RVT-S outperforms ReYOLOv8s by around 9\%, and ReYOLOv8m achieves 80\% of the speed of RVT-B. On the other hand, compared to models that do not deploy RNNs, ReYOLOv8s is, on average, 1.9x slower than other similar-sized models. A similar behavior is also present for Medium models. A deeper look into their design choices should be done to understand this difference. For the case of DTSDNet \cite{dstdnet}, events are converted into two different tensors according to different time-window sizes, a smaller and a bigger one. Those tensors are passed through two parallel CNNs, and the resulting features are fused afterward. By doing so, they can introduce long- and short-term temporal modeling without adopting RNNs, avoiding the latency those cells introduce. AED \cite{aed} achieves a similar effect by creating tensors with long-range temporal information incremented periodically through updates of a FIFO-queue, while HMNET \cite{hmnet} leverages a hierarchical structure with latent memories working at different rates. Furthermore, replacing RNN cells with SSM also led to speed-ups \cite{event-ssm}. However, those models are generally outperformed by RNN-based models, where at a Small scale, ReYOLOV8s has a 2.8\% better mAP than HMNet-L1, with 1.4x fewer parameters and a 3.6\% better performance when compared to S5-ViT-S, which is based on SSM cells. In contrast, at the Medium scale, ReYOLOv8m has a 3.6\% higher performance when compared to DTSDNet-M, with a 1.42x smaller model. ReYOLOv8m also has an average of 5.2\% higher mAP than the SSM-based models S4D-ViT-B and S5-ViT-B.

In conclusion, by integrating recurrent cells and leveraging a robust baseline, the ReYOLOv8 models achieve higher mAP than other models of similar scales. Although they incur some latency penalties compared to non-RNN models, they require fewer parameters. However, the latencies ranging from 9.2ms to 15.5ms are still suitable for real-time operation.

\begingroup
\renewcommand{\arraystretch}{1.0}
\begin{table}[ht!]
\centering

\caption{State-of-the-art performance comparison for the Prophesee's GEN1 dataset.}
\label{tab:gen1_table}
\begin{tabular}{lcccccc}
\cline{2-7}
\multicolumn{1}{c|}{}                                                                                            & \multirow{2}{*}{\textbf{Model}}                                                  & \multirow{2}{*}{\textbf{Network}} & \multirow{2}{*}{\textbf{Parameters}} & \multirow{2}{*}{\textbf{mAP}} & \multicolumn{2}{c|}{\textbf{Runtime}}                                                                                                                                  \\
\multicolumn{1}{c|}{}                                                                                            &                                                                                  &                                   &                                      &                               & \multicolumn{1}{c|}{\textbf{\begin{tabular}[c]{@{}c@{}}v100 \\ GPU\end{tabular}}} & \multicolumn{1}{c|}{\textbf{\begin{tabular}[c]{@{}c@{}}Other\\ GPUs\end{tabular}}} \\ \hline
\multicolumn{1}{|c|}{\multirow{2}{*}{\begin{turn}{+270} Nano \end{turn}}}    & \begin{tabular}[c]{@{}c@{}}RVT-T \\ \cite{rvt} \end{tabular}         & Transformer + RNN                 & \textbf{4.4M}                        & 44.1                          & \textbf{-}                                                                        & \multicolumn{1}{c|}{9.4ms}                                                         \\
\multicolumn{1}{|c|}{}                                                                                           & \textbf{\begin{tabular}[c]{@{}c@{}}ReYOLOv8n \\ (this work)\end{tabular}}        & CNN + RNN                         & 4.7M                                 & \textbf{46.3}                 & \textbf{11.3ms}                                                                   & \multicolumn{1}{c|}{\textbf{9.2ms}}                                                \\ \hline
\multicolumn{1}{|c|}{\multirow{7}{*}{\begin{turn}{+270} Small \end{turn}}}   & \textbf{\begin{tabular}[c]{@{}c@{}}ReYOLOv8s\\ (this work)\end{tabular}}         & CNN + RNN                         & \textbf{8.4M}                        & \textbf{48.3}                 &  13.4ms                                                                 & \multicolumn{1}{c|}{10.4ms}                                                        \\
\multicolumn{1}{|c|}{}                                                                                           & \begin{tabular}[c]{@{}c@{}}DTSDNet-S\\ \cite{dstdnet}\end{tabular}  & Dual Pathway CNN                  & 9.1M                                 & 43.9                          & 7.5ms                                                                             & \multicolumn{1}{c|}{-}                                                             \\
\multicolumn{1}{|c|}{}                                                                                           & \begin{tabular}[c]{@{}c@{}}HMNet-B1\\ \cite{hmnet}\end{tabular}     & HMNet                             & 9.4M                                 & 45.5                          & \textbf{4.6ms}                                                                    & \multicolumn{1}{c|}{\textbf{-}}                                                    \\
\multicolumn{1}{|c|}{}                                                                                           & \begin{tabular}[c]{@{}c@{}}S5-ViT-S\\ \cite{event-ssm}\end{tabular} & Transformer + SSM                 & 9.7M                                 & 46.6                          & -                                                                                 & \multicolumn{1}{c|}{\textbf{7.8ms}}                                                \\
\multicolumn{1}{|c|}{}                                                                                           & \begin{tabular}[c]{@{}c@{}}RVT-S \\ \cite{rvt}\end{tabular}         & Transformer + RNN                 & 9.9M                                 & 46.5                          & \textbf{-}                                                                        & \multicolumn{1}{c|}{9.5ms}                                                         \\
\multicolumn{1}{|c|}{}                                                                                           & \begin{tabular}[c]{@{}c@{}}HMNet-L1\\ \cite{hmnet}\end{tabular}     & HMNet                             & 11.4M                                & 47.0                          & 5.6ms                                                                             & \multicolumn{1}{c|}{-}                                                             \\
\multicolumn{1}{|c|}{}                                                                                           & \begin{tabular}[c]{@{}c@{}}AED\\ \cite{aed}\end{tabular}            & CNN                               & 14.8M                                & 45.4                          & 11.9ms                                                                            & \multicolumn{1}{c|}{-}                                                             \\ \hline
\multicolumn{1}{|l|}{\multirow{10}{*}{\begin{turn}{+270} Medium \end{turn}}} & \begin{tabular}[c]{@{}c@{}}S4D-ViT-B\\ \cite{event-ssm} \end{tabular}         & Transformer + SSM                 & \textbf{16.5M}                                & 46.2                          & -                                                                                 & \multicolumn{1}{c|}{9.4ms}                                                         \\
\multicolumn{1}{|l|}{}                                                                                           & \textbf{\begin{tabular}[c]{@{}c@{}}ReYOLOv8m\\ (this work)\end{tabular}}         & CNN + RNN                         & 18.1M                       & \textbf{49.4}                 &  15.5ms                                                                  & \multicolumn{1}{c|}{12.3ms}                                                        \\
\multicolumn{1}{|l|}{}                                                                                           & \begin{tabular}[c]{@{}c@{}}S5-ViT-B\\ \cite{event-ssm}\end{tabular}          & Transformer + SSM                 & 18.2M                                & 47.7                          & -                                                                                 & \multicolumn{1}{c|}{\textbf{8.2ms}}                                                \\
\multicolumn{1}{|l|}{}                                                                                           & \begin{tabular}[c]{@{}c@{}}RVT-B\\ \cite{rvt}\end{tabular}          & Transformer + RNN                 & 18.5M                                & 47.2                          & \textbf{-}                                                                        & \multicolumn{1}{c|}{10.2ms}                                                        \\
\multicolumn{1}{|l|}{}                                                                                           & \begin{tabular}[c]{@{}c@{}}SAST-CB \\ \cite{sast}\end{tabular}      & Transformer + RNN                 & 18.9M                                & 48.2                          & -                                                                                 & \multicolumn{1}{c|}{-}                                                             \\
\multicolumn{1}{|l|}{}                                                                                           & \begin{tabular}[c]{@{}c@{}}GET\\ \cite{get}\end{tabular}            & Transformer + RNN                 & 21.9M                                & 47.9                          & -                                                                                 & \multicolumn{1}{c|}{16.8ms}                                                        \\
\multicolumn{1}{|l|}{}                                                                                           & \begin{tabular}[c]{@{}c@{}}RED\\ \cite{red_megapixel}\end{tabular} & CNN + RNN                         & 24.1M                                & 40.0                          & -                                                                                 & \multicolumn{1}{c|}{16.7ms}                                                        \\
\multicolumn{1}{|l|}{}                                                                                           & \begin{tabular}[c]{@{}c@{}}DTSDNet-M \\ \cite{dstdnet} \end{tabular} & Dual Pathway CNN                  & 25.7M                                & 47.7                          & 7.5ms                                                                             & \multicolumn{1}{c|}{-}                                                             \\
\multicolumn{1}{|l|}{}                                                                                           & \begin{tabular}[c]{@{}c@{}}HMNet-B3\\ \cite{hmnet}\end{tabular}     & HMNet                             & 29.8M                                & 45.2                          & \textbf{7.0ms}                                                                    & \multicolumn{1}{c|}{\textbf{-}}                                                    \\
\multicolumn{1}{|l|}{}                                                                                           & \begin{tabular}[c]{@{}c@{}}HMNet-L3\\ \cite{hmnet}\end{tabular}     & HMNet                             & 33.2M                                & 47.1                          & 7.9ms                                                                  & \multicolumn{1}{c|}{-}                                                             \\ \hline
                                                                                                                 & \multicolumn{1}{l}{}                                                             & \multicolumn{1}{l}{}              & \multicolumn{1}{l}{}                 & \multicolumn{1}{l}{}          &                                                                                   & \multicolumn{1}{l}{}                                                              
\end{tabular}
\end{table}

\endgroup

\section{Conclusion}
\label{sec:Conclusion}

In this work, we proposed a framework for event-based object detection by modifying the frame-based detector YOLOv8 to include recurrent cells, resulting in the new Recurrent YOLOv8 (ReYOLOv8) model. Additionally, we introduced novel event encoding and event-specific data augmentation techniques. The proposed event encoding, called Volume of Ternary Event Images (VTEI), leverages the spatio-temporal dynamics of event streams while minimizing processing time, memory requirements, and bandwidth. VTEI was shown to be 2.35 times faster than alternatives when processing peak event loads, requiring only 3 bytes per non-zero and a peak bandwidth of 9.54 MB/s.

Three scale-based variations of ReYOLOv8—Nano, Small, and Medium—were evaluated using two large-scale, real-world datasets: PEDRo and GEN1. To address polarity imbalances in event streams, we proposed a data augmentation technique that randomly suppresses specific polarities, leading to an average mAP improvement of 0.7\% for GEN1 and 4.5\% for PEDRo. Compared to existing literature, the ReYOLOv8 models demonstrated a 9\% to 18\% improvement in mAP for PEDRo, while requiring 8.8x fewer parameters and being 1.67x faster on average. For GEN1, the models presented outstanding performance across all the similar scales, achieving mAP improvements of approximately 5\%, 2.8\%, and 2.5\% over the runner-ups up to 5M, 15M, and 45M parameters, respectively, with an average parameter reduction of 20.8\%. The inference speeds, ranging from 9.2ms to 15.5ms, make these models suitable for real-time operation.

Future work will extend the Random Suppression Polarity analysis to other event encodings. Additionally, there is a need to create comprehensive benchmarks to better understand the system-level impacts of jointly designing event encodings and detection models, as this remains underexplored. Finally, we intend to broaden the framework evaluation to include other datasets available in the literature.

\section*{Conflict of Interest Statement}
The authors declare that the research was conducted in the absence of any commercial or financial relationships that could be considered a potential conflict of interest.



\section*{Funding}

This work has been funded by King Abdullah University of Science and Technology CRG program under grant number: URF/1/4704-01-01.

\section*{Acknowledgments}

We express our sincere gratitude to the KAUST Supercomputing Laboratory for granting us access to its GPU clusters. Additionally, we are deeply thankful to Mathias Gehrig from the Robotics and Perception Group at the University of Zurich for his invaluable insights and information on the implementation aspects of recurrent-convolutional architectures and managing large-scale event datasets.

\section*{Data Availability Statement}
The datasets analyzed for this study can be found in the PEDRo-Event-Based-Dataset [\url{https://github.com/SSIGPRO/PEDRo-Event-Based-Dataset}] and Prophesee Automotive Detection Dataset [\url{https://www.prophesee.ai/2020/01/24/prophesee-gen1-automotive-detection-dataset/}] repositories.

\end{document}